\documentclass[lettersize,journal]{IEEEtran}

\usepackage{amsmath,amsfonts,amssymb,mathtools}

\usepackage{graphicx}
\usepackage{array}
\usepackage[caption=false,font=normalsize,labelfont=sf,textfont=sf]{subfig}
\usepackage{booktabs}
\usepackage{multirow}
\usepackage{tabularx}

\usepackage{algorithm}
\usepackage[noend]{algpseudocode}

\usepackage{textcomp}
\usepackage{stfloats}
\usepackage{etoc}
\usepackage{url}
\usepackage{cite}
\usepackage[table]{xcolor}
\usepackage{siunitx}      
\usepackage{xspace}
\usepackage{tcolorbox}
\newcommand{\eg}{e.g.,\xspace}

\usepackage[hidelinks]{hyperref}
\usepackage[utf8]{inputenc}
\usepackage{enumitem}
\usepackage{capt-of}
\usepackage{xcolor}
\definecolor{RareFlowColor}{HTML}{4E95D9}

\begin{document}

\title{Semantic-Guided Cross-Sensor Super Resolution of Remote Sensing Images: A Gated Dual Conditioning Flow Matching Model
}

\author{%
Forouzan~Fallah\IEEEauthorrefmark{1},
Wenwen~Li\IEEEauthorrefmark{2}\IEEEauthorrefmark{3},
Chia-Yu~Hsu\IEEEauthorrefmark{2},
Hyunho~Lee\IEEEauthorrefmark{2},
Anna Liljedahl\IEEEauthorrefmark{4},
Yezhou~Yang\IEEEauthorrefmark{1}%
\thanks{\IEEEauthorrefmark{1}School of Computing and Augmented Intelligence, Arizona State University.}%
\thanks{\IEEEauthorrefmark{2}School of Geographical Sciences and Urban Planning, Arizona State University.}%
\thanks{\IEEEauthorrefmark{4}Woodwell Climate Research Center}%
\thanks{\IEEEauthorrefmark{3}Corresponding author. (e-mail: wenwen@asu.edu)}%
}

\markboth{Journal of \LaTeX\ Class Files,~Vol.~14, No.~8, August~2021}%
{Shell \MakeLowercase{\textit{et al.}}: A Sample Article Using IEEEtran.cls for IEEE Journals}

\maketitle
\begin{abstract}
High spatial resolution satellite imagery is critical for monitoring fine-scale Earth surface processes, but is often limited by cost and revisit time. This work studies cross-sensor super-resolution (SR) to reduce this gap by translating 10 m Sentinel-2 imagery into 2 m Maxar-like imagery in a data-scarce, domain-shifted setting, with a focus on rare geomorphic features such as retrogressive thaw slumps (RTS). We propose RareFlow, a semantic-guided generative AI framework for cross-sensor super-resolution based on a flow-matching formulation, designed to produce visually plausible and physically reliable high-resolution images. RareFlow uses dual conditioning to guide the generation process: (1) a gated ControlNet that preserves scene geometry from low-resolution (LR) input, and (2) text-based semantic guidance that injects contextual information when the target phenomenon is rare. To ensure high-fidelity outputs, we introduce a multifaceted loss function that anchors the output to the high-resolution (HR) ground truth by jointly enforcing frequency alignment, perceptual similarity, and color consistency. RareFlow’s performance is systematically evaluated on a newly curated benchmark of multi-sensor satellite imagery for rare Earth feature detection, and its generalizability is demonstrated on two public remote sensing benchmarks, SEN2NAIP and BreizhSR. Human evaluation with domain experts is also conducted to further verify RareFlow’s effectiveness in generating high-fidelity super-resolved images for scientific analysis.
Project Page: \href{https://rareflow.github.io}{RareFlow.github.io}
\end{abstract}

\etocdepthtag.toc{main}
\section{Introduction}
\label{sec:intro}

Monitoring highly variable, fine-scale surface processes such as retrogressive thaw slump (RTS) activity, coastal erosion, and slope failure across broad regions requires Earth observation (EO) imagery with both fine spatial detail and frequent revisit times\cite{Barth2025,caldareri2024shoreline,casagli2023landslide}. These requirements are rarely met by a single sensing platform. Fine spatio-temporal monitoring of the land surface further enables a refined understanding of the erosional processes and controlling factors, for example, the role of weather anomalies such as wind, rainfall or air temperature on slope failure initiation, expansion, and stabilization. Public satellite missions such as European Space Agency (ESA)'s Sentinel-2 provide near-global coverage and dense temporal sampling, but their spatial resolution is often insufficient to resolve fine-scale morphology relevant to many EO applications \cite{gascon2017sentinel2calib,ESA_Sentinel2}. In contrast, very-high-resolution (VHR) commercial imagery such as Maxar (now Vantor) Worldview imagery provides the spatial detail needed, with centimeter level spatial resolution, for feature-level interpretation, but it is costly, less accessible, and not consistently available for repeated large-scale monitoring \cite{Maxar_WorldView3,turner2015free}. This trade-off creates a sensing gap between imagery that is frequently available and imagery that supports reliable fine-scale analysis.

Cross-sensor super-resolution (SR) addresses this gap by estimating high-resolution (HR) content in a target sensor domain from a low-resolution (LR) input acquired by another sensor \cite{qiu2023cross,li2023convformersr}. In this setting, the goal is not merely to sharpen the LR image. The output must recover fine-scale content in a target domain with differing spatial, radiometric, and textural characteristics \cite{qiu2023cross,latte2020planetscope}. This distinction is important in remote sensing, where the best available HR reference often comes from a different sensor rather than from the same instrument at a finer resolution. The task is therefore better viewed as target-domain reconstruction under inter-sensor variability than as conventional same-sensor SR \cite{qi2026advancing}.

This problem is especially challenging in remote sensing because perceptually sharp imagery alone is not sufficient. Reconstructed detail must remain scientifically interpretable and consistent with the observed scene structure and the target-sensor's characteristics \cite{aybar2026radiometrically}. For example, when mapping a mass-wasting feature like an RTS, the scientist would preferably be able to outline the main morphological parts of the feature such as the headwall, slump floor, and mud flow. A useful cross-sensor SR method must therefore satisfy three requirements simultaneously. First, it must preserve structural fidelity to the LR input image so that generated details remain grounded in the measured scene layout. Second, the reconstruction must introduce plausible fine-scale details for environmental features of interest, even when these features are rare and only weakly expressed in coarse imagery. Third, the output must maintain cross-sensor consistency for scientific use, including compatibility with the radiometric, spatial, and textural characteristics of the target sensor. In practice, these requirements are difficult to satisfy simultaneously because preserving source-grounded structure can limit inferred detail, while rare-feature plausibility and target-sensor consistency often rely on information that is only weakly supported by the LR input image under cross-sensor heterogeneity. The central challenge is therefore to control how strongly generative priors (i.e., learned patterns from training data) should influence reconstruction when the input evidence is blurred, ambiguous, or out of distribution \cite{wu2024seesr,le2024detecting}.

\begin{figure}[h]
    \centering
    \includegraphics[width=\linewidth]{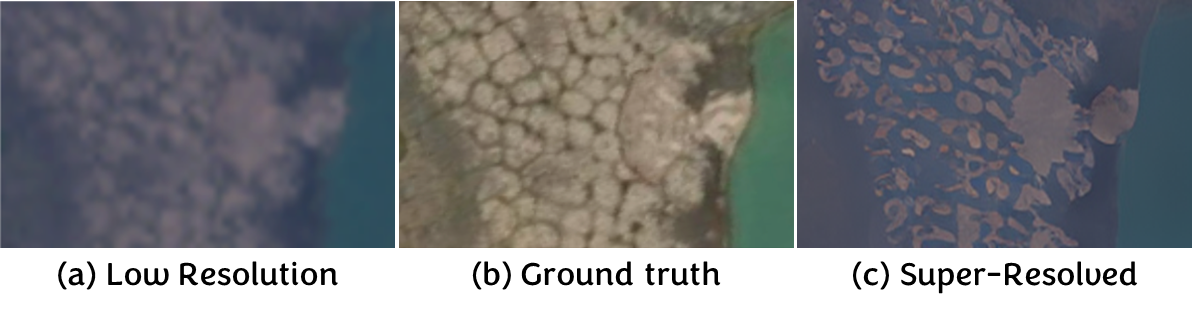}
    \caption{An example of a failure case for the state-of-the-art model’s super-resolved image is shown in (c). While it appears sharper and more plausible than the low-resolution image in (a), it fails to capture the true morphology of ice-wedge polygon landscape features that are shown in the high-resolution image (b).}
    \label{fig:example}
\end{figure}

Prior SR methods remain limited in addressing this problem. Traditional interpolation and early deep SR models often oversmooth reconstructions and suppress fine spatial detail \cite{ledig2017photo,blau2018perception}. More recent generative and diffusion-based approaches improve perceptual sharpness and texture realism, but they can also hallucinate unsupported structures when the input does not sufficiently constrain the reconstruction \cite{saharia2022image,wang2024exploiting}. In remote sensing, this failure mode is especially problematic because plausible-looking detail may not correspond to scientifically meaningful scene content \cite{aybar2026radiometrically}. An example is illustrated in Fig. \ref{fig:example}, where the ice-wedge polygon surface feature is recognizable to expert earth scientists in both the low- and high-resolution images, while the feature has been distorted beyond recognition in the super-resolved image. Many existing remote-sensing SR methods do not address, in a unified way, the combined challenges of rare features, cross-sensor domain shift, and scientific interpretation, and they provide little explicit control over prior-driven generation when observational evidence is limited \cite{qi2026advancing,qiu2023cross,lu2025multi}.

To address these limitations, we propose RareFlow, a semantically guided, spatially controlled cross-sensor SR model motivated by the needs of the scientific analysis of remote sensing imagery under inter-sensor variability, limited data, and domain shift. RareFlow builds on a strong generative AI prior (i.e., Stable Diffusion) \cite{esser2024scaling,Peebles2023DiT} by constraining prior-driven synthesis with spatial conditioning to ensure that reconstructed details remain grounded in the LR input and consistent with the target sensor. It achieves this by introducing four complementary components. First, a controllable SR image generation pathway conditions the Stable Diffusion model on the LR input image to preserve scene structure and constrain where fine detail can plausibly appear \cite{zhang2023adding,zavadski2024controlnet}. Second, semantic text guidance provides task-relevant context for plausible fine-scale reconstruction \cite{wu2024seesr}. For environmental monitoring of rare Earth features, this guidance can encode feature-specific morphology. More broadly, it can serve as a caption-level semantic prior for SR. Third, a learnable, uncertainty-gated mechanism adaptively modulates the LR input with semantic-guided generative features to reduce overreliance on prior-driven synthesis and better handle out-of-distribution inputs \cite{zhang2025uncertainty}. Finally, a consistency-guided training objective encourages reconstructions to remain aligned with the radiometric and textural characteristics of the target sensor domain. Together, these controls are designed to limit unsupported synthesis while preserving scene structure, rare-feature plausibility, and cross-sensor consistency.   

We evaluate RareFlow on a curated cross-sensor benchmark of rare Earth features (e.g., retrogressive thaw slumps, which are important indicators of climate change impacts). Experimental results show that RareFlow improves the balance among structural fidelity, perceptual quality, and target sensor consistency, while the expert evaluation further supports its value for interpreting rare geomorphic features. RareFlow’s generalizability is further validated on two additional cross-sensor SR benchmark datasets based on remote-sensing imagery.

The remainder of this paper reviews related work, describes RareFlow and its training objective, presents experiments on paired and controlled settings, and discusses uncertainty, failure cases, and implications of the proposed method for scientific analysis.

\section{Related Work}
\label{sec:related-works}

Remote-sensing SR aims to improve the practical utility of widely available medium-resolution imagery for applications that require finer spatial detail. In EO, however, SR is not only a visual enhancement task, it also needs to ensure that the reconstructed outputs must remain interpretable with respect to the observed scene and suitable for scientific analysis. Much of the existing literature addresses this problem under matched-sensor or synthetically degraded settings, where the HR target is treated as a sharper version of the LR input. We therefore begin with this line of work, then turn to real-world cross-sensor SR and to generative methods that emphasize structural grounding, semantic guidance, adaptive control of prior-driven synthesis, and target-sensor consistency.

\paragraph{Remote Sensing Super-Resolution Under Matched-Sensor or Synthetic Degradation Settings}

Most remote-sensing SR studies are built on paired LR-HR data, where the degradation from HR to LR is a known or synthetically defined \cite{liu2021research,wang2022review}. Under this formulation, the HR target is a sharper version of the same scene rather than an observation from a different sensor domain. Early deep SR models such as SRCNN \cite{dong2015image} established the supervised paradigm that an end-to-end learned mapping from LR to HR could outperform hand-crafted pipelines. In remote sensing, DSen2 \cite{lanaras2018super} extended it to Sentinel-2 by learning to super-resolve 20m or 60m bands to 10m from synthetically downsampled same-sensor data. Later transformer-based models such as SwinIR \cite{liang2021swinir} improved restoration quality under the same assumptions with increased model capacity and context modeling that captures longer-range dependencies. These methods are effective when LR-HR pairs are well aligned, the scale-transfer assumption holds, and evaluation is based on fidelity-oriented measures. However, they do not directly address sensor mismatch, target-domain reconstruction, or scientific consistency across sensors.

\paragraph{Cross-Sensor and Real-World Super-Resolution Under Sensor Mismatch}

Recent work has moved remote-sensing SR beyond synthetic bicubic degradation toward settings in which the LR observation and HR reference come from different sensors and often from different times. Benchmark efforts such as MuS2 \cite{kowaleczko2023real} and SEN2NAIP \cite{aybar2024sen2naip} make this shift explicit. MuS2 frames Sentinel-2 multi-image SR against WorldView-2 references under real cross-sensor and cross-temporal discrepancy. SEN2NAIP similarly builds a large-scale Sentinel-2 to NAIP benchmark and shows that scalable training requires explicit harmonization of blur, spectral response, and noise rather than synthetic downsampling alone. Together, these datasets show that real-world EO SR is fundamentally a sensor-mismatch problem.

Methods in this setting often add extra structure rather than relying on direct LR-to-HR regression. MISR-S2 \cite{okabayashi2024cross} addresses Sentinel-2 to SPOT-6 SR from irregularly sampled time series with a time-equivariant fusion module, showing that temporal modeling helps when the LR-HR gap is large. Ref-Diff \cite{dong2024building} instead uses an HR reference from another time together with change priors to regulate texture transfer in a diffusion-based framework. Michel \textit{et al.} further show that geometric and radiometric distortion are central factors in cross-sensor SR rather than minor nuisances \cite{michel2025revisiting}. Together, these studies show that real-world cross-sensor SR often requires extra structure to cope with temporal discrepancy, sensor mismatch, and imperfect alignment. However, they still leave open the setting studied here: direct target-domain reconstruction under strong sensor mismatch, where generated detail must remain structurally grounded, compatible with the target sensor, and scientifically interpretable.

\paragraph{Grounded Generative Super-Resolution Combining Spatial and Semantic Controls}

Recent generative SR increasingly treats reconstruction as a problem of controlled generation rather than unconstrained texture synthesis \cite{gandikota2024text,moser2024diffusion}. The main challenge is not only how to produce sharper images, but how to generate fine-scale detail while remaining grounded in the LR observation and scientifically usable under sensor mismatch. In this setting, the most relevant literature centers on four closely related themes that RareFlow later combines: structural grounding, semantic guidance, adaptive control of prior-driven synthesis, and target-sensor consistency.

For structural grounding, ControlNet \cite{zhang2023adding} shows that a pretrained diffusion model can follow explicit spatial controls through a lightweight control branch. In SR, this idea has been adapted to preserve scene layout while leveraging generative priors (i.e., the learned distribution of plausible HR patterns encoded by the generative model) \cite{wang2024exploiting, yang2024pixelaware, wan2024controlsr}. Along similar lines, ZoomLDM \cite{ZoomLDM} conditions latent diffusion on scale and self-supervised features to support multi-scale generation, while SAMSR \cite{samsr} introduces segmentation or structural priors to sharpen boundaries and improve local consistency. Together, these works show that structural conditioning is important for limiting unsupported detail and keeping generated content tied to the observed scene geometry, especially under sensor mismatch.

\begin{figure*}[t]
    \centering
    \includegraphics[width=1\linewidth]{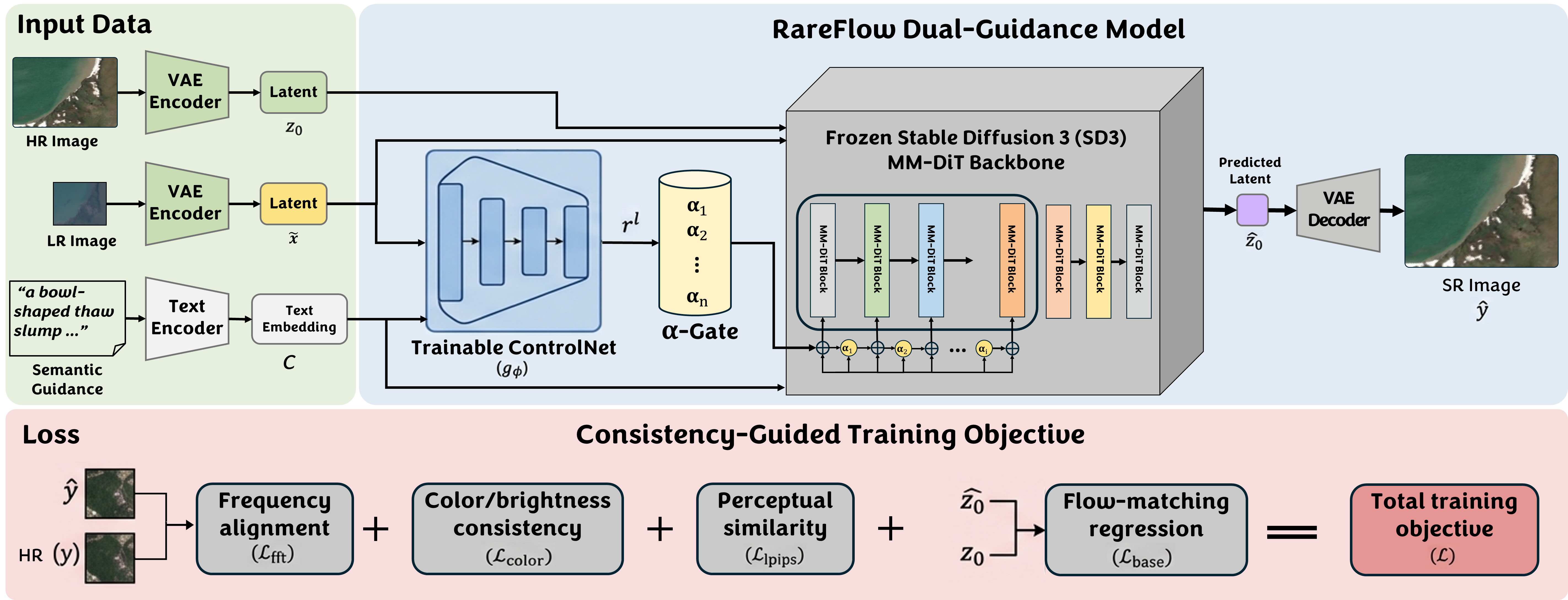}
    \caption{Overview of the RareFlow framework. The top panel illustrates the training workflow, where a frozen SD3 backbone is conditioned on both LR latents and semantic guidance via a trainable ControlNet and an $\alpha$-Gate mechanism. HR ground truth ($z_0$) is utilized during training to calculate the multi-component Consistency-Guided Training Objective (bottom panel), which includes frequency alignment, color consistency, perceptual similarity, and flow-matching regression to update the ControlNet parameters. The HR pathway and the consistency-guided objective shown in the bottom panel are used only during training; at inference time, RareFlow keeps only the LR-conditioned and text-conditioned forward sampling path to generate $\hat{y}$.
    }
    \label{fig:method1}
\end{figure*}

Structure alone, however, is often insufficient in ill-posed SR settings, where LR images constrain where detail may appear but not what fine-scale detail should be. Semantic priors help narrow the space of plausible reconstructions and reduce implausible textures \cite{qu2024xpsr,xiao2024semantic,sftgan}. SeeSR \cite{wu2024seesr}, for example, uses degradation-aware semantic prompts to reduce semantically incorrect detail under severe degradation, while, in remote sensing, studies show that combining semantic guidance with diffusion improves large-factor SR \cite{SGDM}. These studies highlight the value of semantic guidance, when rare content under cross-sensor degradation might otherwise be replaced by more common patterns from the training distribution.

Once structural and semantic guidance are introduced, the next question is how strongly the generative prior should influence the reconstruction. Recent work suggests that this influence should be adaptive rather than fixed. StableSR \cite{wang2024exploiting} illustrates controllable fidelity-quality trade-offs on top of a frozen diffusion prior, and AdcSR \cite{chen2025adversarial} shows that strong distilled diffusion priors can retain perceptual quality in real-world SR. This perspective connects directly to uncertainty-aware control. Zhang \textit{et al.} use uncertainty estimates to modulate diffusion behavior itself, allowing more generative freedom in ambiguous regions while keeping confident regions closer to LR evidence \cite{zhang2025uncertainty}. In remote sensing, OpenSR (LDSR-S2) \cite{donike2025trustworthy} similarly brings uncertainty into an EO latent-diffusion setting framed around trustworthy reconstruction. Collectively, these studies suggest that prior-driven generation should be selective and evidence-aware rather than uniformly applied.

Taken together, this literature increasingly recognizes that cross-sensor super-resolution for EO data requires grounded generation, semantic context, adaptive control of prior-driven synthesis, and explicit target-domain consistency. Yet these elements are still rarely unified in a single framework for direct target-domain reconstruction under strong sensor mismatch, leaving a gap for methods designed to preserve structure, regulate prior influence under ambiguity, and maintain target-sensor compatibility at once.

\section{Proposed Methodology: RareFlow}
\label{sec:method}

Fig.~\ref{fig:method1} summarizes the RareFlow framework. In remote-sensing single-image SR, the main challenge is to recover useful high-frequency detail without creating structures that are not supported by the LR observation. In many cases, the LR input alone is too ambiguous to determine reliable fine structure. RareFlow addresses this problem by combining two complementary sources of guidance: (i) semantic guidance from descriptive text captions through a frozen SD3 latent diffusion backbone \cite{Esser2024SD3,Peebles2023DiT}, and (ii) observation guidance from an LR-conditioned ControlNet branch \cite{zhang2023adding,zavadski2024controlnet}. To balance these two sources, RareFlow learns block-wise scalar gates that control how strongly the LR-conditioned residuals are injected into the frozen backbone. As a result, the model can limit over-reliance on LR evidence while preserving the semantic prior provided by the text-conditioned backbone.

\subsection{Problem setup and latent-space backbone}
\label{subsec:setup}
Let $x$ be an LR image and $y$ its corresponding HR target. Let $x^\uparrow$ denote the LR image after the fixed spatial alignment used in the pipeline. A frozen variational autoencoder (VAE), with encoder $\mathcal{E}$ and decoder $\mathcal{D}$, maps both images to latent space:
\begin{equation}
z_0 = \mathcal{E}(y), \qquad \tilde{x} = \mathcal{E}(x^\uparrow).
\end{equation}

RareFlow builds on a frozen SD3 MM-DiT backbone $f_\theta$ \cite{Peebles2023DiT,Esser2024SD3} (See Fig. \ref{fig:method}(b) for its inner structure). Following the SD3 Flow-Matching formulation, a noisy latent is formed at diffusion step $t$ as
\begin{equation}
z_t = (1-\sigma_t)\,z_0 + \sigma_t\,\epsilon,
\qquad \epsilon \sim \mathcal{N}(0,\mathbf{I}),
\label{eq:raref-low-noise}
\end{equation}
where $\{\sigma_t\}_{t=0}^{N}$ is the discrete noise schedule of the SD3 Euler solver. The denoising network receives the noisy latent $z_t$, the time step $t$, and the text conditioning $c$ used by the SD3 interface. Starting from Gaussian noise, inference follows the standard SD3 sampling schedule, and the final latent is decoded by $\mathcal{D}$.

\subsection{Semantic guidance from descriptive captions}
\label{subsec:semantic}

RareFlow uses descriptive text captions as semantic conditioning for the frozen SD3 backbone. The caption embedding $c$ provides scene-level guidance that is often ambiguous in LR image alone, especially when the input is blurred, aliased, or lacks reliable fine structure. In remote-sensing SR, this semantic guidance helps the model favor structures that are consistent with the scene content instead of relying only on local evidence from the LR observation.

The text caption does not provide exact target details. Instead, it gives a high-level semantic prior that guides the denoising trajectory toward scene-consistent reconstructions. In RareFlow, this semantic prior is carried by the frozen SD3 backbone throughout sampling, while the LR observation is introduced through the ControlNet path described next.

\subsection{LR-conditioned ControlNet with learned gates}
\label{subsec:controlnet}

\begin{figure*}[t]
    \centering
    \includegraphics[width=.8\textwidth]{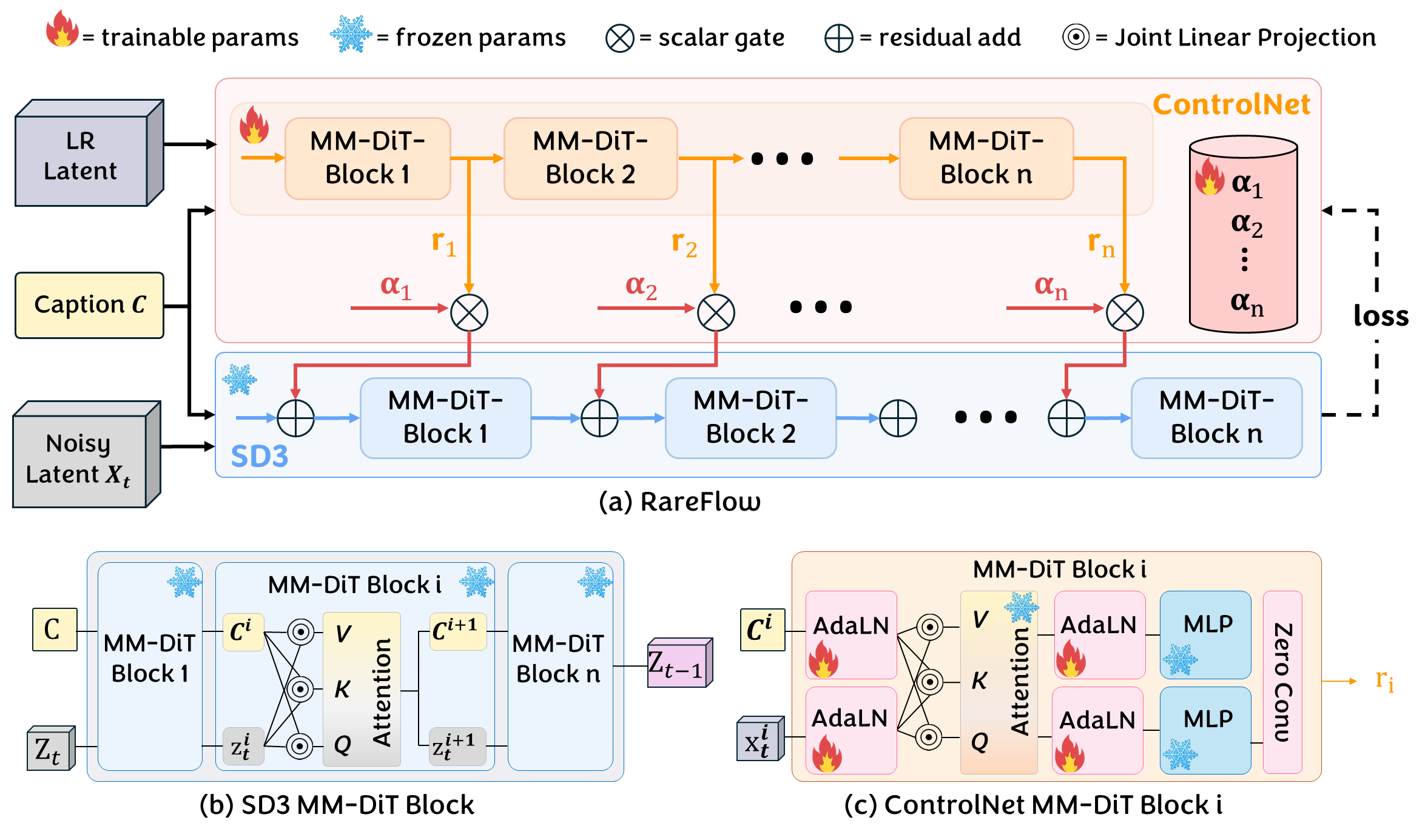}
    \caption{
    \textbf{Technical architecture of RareFlow.}
    \textbf{(a)} RareFlow uses dual conditioning: descriptive text captions condition the frozen SD3 backbone (\textcolor{blue}{blue}), while the LR latent is injected through a trainable ControlNet branch (\textcolor{orange}{orange}). The ControlNet produces block-wise residual features $r^l$, which are scaled by learned scalar gates $\alpha^l \in [0,1]$ before being added to the corresponding frozen backbone features, i.e.,
    $F^l \leftarrow F^l + \alpha^l r^l$.
    This design balances semantic guidance from text with observational guidance from the LR input, helping reduce over-reliance on LR input.
    \textbf{(b)} Internal structure of an SD3 MM-DiT block.
    \textbf{(c)} Internal structure of a ControlNet MM-DiT block that produces the residual feature $r^i$; the residual is scaled by the learned gate $\alpha^i$ before injection into the frozen backbone.
        }
    \label{fig:method}
\end{figure*}

To inject observation-specific information, RareFlow adds a trainable ControlNet branch $g_\phi$ \cite{zhang2023adding,zavadski2024controlnet} conditioned on the LR latent $\tilde{x}$. The ControlNet produces residual feature maps $\{r^l\}_{l\in\mathcal{L}}$ aligned with a set of selected transformer blocks $\mathcal{L}$. Let $h^l$ denote the hidden feature map at block $l$. RareFlow injects the ControlNet signal as
\begin{equation}
\tilde{r}^l = \alpha^l\,r^l,
\qquad
h^l \leftarrow h^l + \tilde{r}^l,
\label{eq:raref-low-inject}
\end{equation}
where $\alpha^l \in [0,1]$ is the learned scalar gate for block $l$. Fig.~\ref{fig:method} illustrates the detailed architecture of RareFlow. Panel (a) shows the dual-conditioned generation process based on a trainable ControlNet, while panel (c) presents the internal structure of a ControlNet MM-DiT block.

The role of the gates is to regulate how strongly LR-conditioned information enters the frozen backbone. A larger $\alpha^l$ gives more influence to the ControlNet residual at block $l$, while a smaller $\alpha^l$ weakens the LR control signal and leaves more of the denoising process to the text-conditioned SD3 prior. In this way, RareFlow balances observation guidance and semantic guidance rather than forcing the reconstruction to rely only on the LR input.

The set of gates $\{\alpha^l\}_{l\in\mathcal{L}}$ are model parameters learned jointly with the ControlNet. After training, these gates are fixed and used directly at test time.

The full RareFlow predictor is written as
\begin{equation}
\hat{z}_0 = \mathcal{M}(z_t,t,c,\tilde{x};\theta,\phi,\alpha),
\label{eq:raref-low-model}
\end{equation}
where $\theta$ denotes the frozen SD3 parameters, $\phi$ the trainable ControlNet parameters, and $\alpha=\{\alpha^l\}_{l\in\mathcal{L}}$ the learned block-wise gates. The reconstructed SR image is then
\begin{equation}
\hat{y} = \mathcal{D}(\hat{z}_0).
\label{eq:raref-low-yhat}
\end{equation}

\subsection{Consistency-guided training objective}
\label{subsec:train}
During training, only the ControlNet parameters $\phi$ and the gates $\alpha$ are updated, while $(\mathcal{E},\mathcal{D})$ and $f_\theta$ remain frozen. For a sampled pair $(x,y)$, a diffusion step $t$, and Gaussian noise $\epsilon$, RareFlow predicts $\hat{z}_0$ using Eq.~\eqref{eq:raref-low-model} and decodes $\hat{y}$ using Eq.~\eqref{eq:raref-low-yhat}.

The base loss is a preconditioned Flow-Matching regression term \cite{lipman2022flow}:
\begin{equation}
\mathcal{L}_{\mathrm{base}} =
\mathbb{E}_{(x,y),\,t,\,\epsilon}
\left[
\left\|
\omega(\sigma_t)\,(\hat{z}_0-z_0)
\right\|_2^2
\right],
\label{eq:raref-low-base}
\end{equation}
where $\omega(\sigma_t)$ is the schedule-dependent weight.

The base term is complemented by three image-level consistency terms.

\paragraph{Frequency alignment.}
\begin{figure}
    \centering
    \includegraphics[width=1\linewidth]{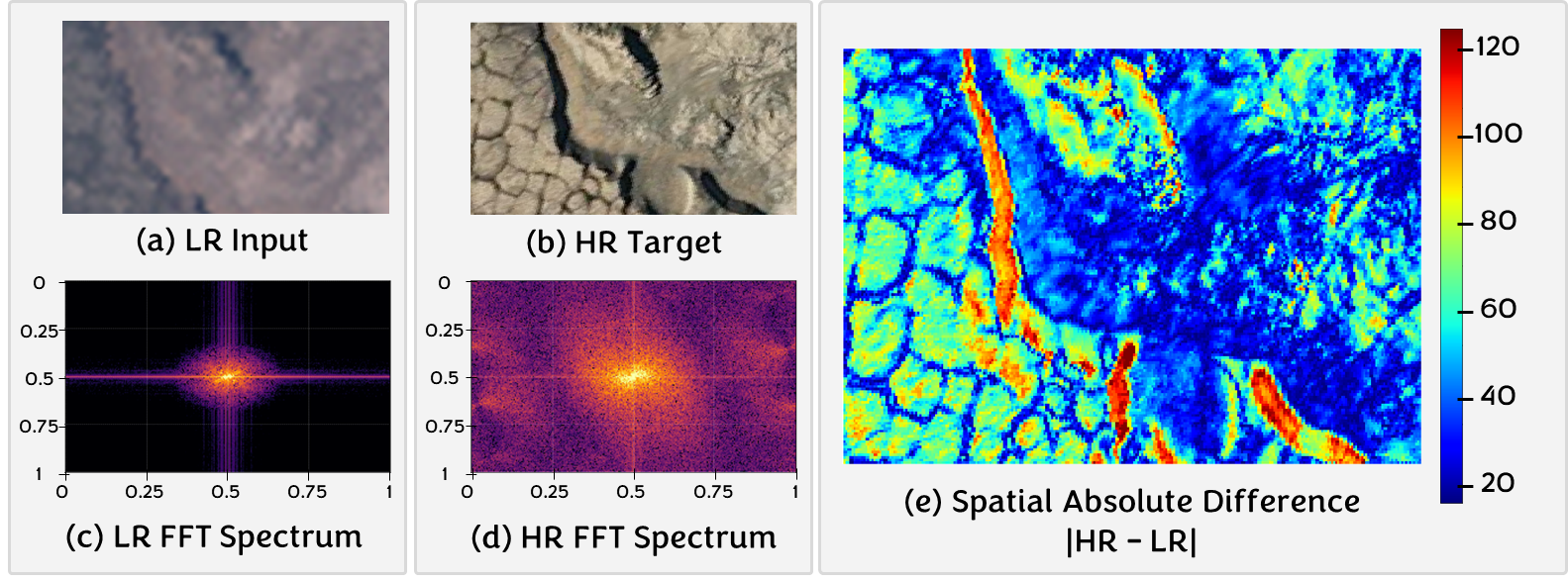}
    \caption{
Comparison of the LR input and HR target in their frequency content. (a) and (b) show the LR input and corresponding HR target. (c) and (d) present the log-scaled 2D FFT magnitude spectra, revealing that the HR target contains substantially stronger high-frequency components than the LR input. This spectral discrepancy motivates the use of an FFT-based loss, which encourages the reconstruction to match the frequency characteristics of the HR target and recover fine details missing from the LR input.  (e) shows the absolute difference between the LR and HR images for visual reference.
    }
    \label{fig:freq}
\end{figure}
To better preserve mid- and high-frequency content (See Fig. \ref{fig:freq}), RareFlow aligns the spectral magnitudes of the prediction and the HR target \cite{Jiang2021FFL,Fuoli2021FourierSR}:
\begin{equation}
\mathcal{L}_{\mathrm{fft}} =
\left\|
W \odot
\Big(
\left|\mathcal{F}(\hat{y})\right| -
\left|\mathcal{F}(y)\right|
\Big)
\right\|_1,
\label{eq:raref-low-fft}
\end{equation}
where $\mathcal{F}$ is the 2D Fourier transform and $W(\rho)\propto \rho^\gamma$ gives more weight to mid- and high-frequency bands.

\paragraph{Coarse color and brightness consistency.}
To preserve low-frequency appearance, RareFlow enforces consistency with the HR target in the CIELAB color space, where $L^\ast$ denotes lightness and $a^\ast,b^\ast$ denote chromatic components. Since this representation separates brightness from color and is better aligned with perceptual color differences, comparing blurred CIELAB images helps capture coarse color and brightness consistency. Let
\begin{equation}
\mathrm{Lab}_b(u) = G_b(\mathrm{Lab}(u)),
\end{equation}
where $\mathrm{Lab}(\cdot)$ converts an RGB image to CIELAB and $G_b$ applies a Gaussian blur to each channel. Let $\mu(\cdot)$ and $\mathrm{std}(\cdot)$ denote the channel-wise spatial mean and standard deviation. The color-consistency loss is
\begin{equation}
\begin{aligned}
\mathcal{L}_{\mathrm{color}} &=
\left\|
\mathrm{Lab}_b(\hat{y}) - \mathrm{Lab}_b(y)
\right\|_1 \\
&\quad+
\left\|
\mu(\mathrm{Lab}_b(\hat{y})) - \mu(\mathrm{Lab}_b(y))
\right\|_1 \\
&\quad+
\left\|
\mathrm{std}(\mathrm{Lab}_b(\hat{y})) - \mathrm{std}(\mathrm{Lab}_b(y))
\right\|_1 .
\end{aligned}
\label{eq:raref-low-color}
\end{equation}

\paragraph{Perceptual similarity.}
Finally, RareFlow uses the LPIPS loss (AlexNet) \cite{lpips}:
\begin{equation}
\mathcal{L}_{\mathrm{lpips}} =
\mathrm{LPIPS}(\hat{y},y).
\label{eq:raref-low-lpips}
\end{equation}
This term complements the latent, frequency, and color losses by encouraging the reconstructed image to remain perceptually close to the HR target at the local-structure level.

\paragraph{Total objective.}
The final training objective is
\begin{equation}
\mathcal{L} =
\mathcal{L}_{\mathrm{base}}
+ \lambda_{\mathrm{fft}}\,\mathcal{L}_{\mathrm{fft}}
+ \lambda_{\mathrm{color}}\,\mathcal{L}_{\mathrm{color}}
+ \lambda_{\mathrm{lpips}}\,\mathcal{L}_{\mathrm{lpips}}.
\label{eq:raref-low-total}
\end{equation}
This objective keeps the latent prediction close to the HR target while also encouraging the final image to match the target in frequency content, coarse appearance, and perceptual quality.

\subsection{Training and inference procedure}
\label{subsec:train-infer}

Rareflow follows slightly different training and inference schemes. During training (refer to Fig.~\ref{fig:method1}), the model has access to the LR input $x$, the HR target $y$, and the descriptive text caption. The aligned LR image $x^\uparrow$ and the HR target $y$ are first encoded by the frozen VAE into latents $\tilde{x}=\mathcal{E}(x^\uparrow)$ and $z_0=\mathcal{E}(y)$. A noisy latent $z_t$ is then constructed from $z_0$ using the SD3 flow-matching noise schedule, as defined in Eq.~\eqref{eq:raref-low-noise}. The frozen SD3 backbone receives $(z_t,t,c)$, while the trainable ControlNet, conditioned on $\tilde{x}$, produces residual features that are injected into selected MM-DiT blocks after scaling by the learned gates $\alpha^l$, as in Eq.~\eqref{eq:raref-low-inject}. This produces the latent prediction $\hat{z}_0$, which is decoded into the SR image $\hat{y}$.

The HR target is used only during training. Specifically, $z_0$ is required to form the noisy training latent $z_t$, and $y$ is required to compute the consistency-guided objective in Eq.~\eqref{eq:raref-low-total}. RareFlow is trained with a hybrid objective consisting of one latent-space regression term and three image-space consistency terms. The flow-matching base loss supervises the predicted clean latent $\hat{z}_0$ against the HR latent $z_0=\mathcal{E}(y)$, while the FFT, color, and LPIPS losses are computed in image space between the decoded prediction $\hat{y}=\mathcal{D}(\hat{z}_0)$ and the HR target $y$.
Gradients from this objective update only the ControlNet parameters $\phi$ and the gates $\alpha$, while the VAE and SD3 backbone remain frozen.

During inference, the HR target $y$ is not available, so the HR branch and all loss terms are removed. RareFlow uses only the LR image $x$ and the text caption. The LR image is encoded into $\tilde{x}$, and sampling starts from Gaussian noise in latent space. The model then follows the standard SD3 flow-matching sampling procedure: at each step, the frozen SD3 backbone is guided by the text embedding $c$, and the learned ControlNet residuals, modulated by the fixed gates $\alpha$, inject LR-specific information into the denoising trajectory. After the final sampling step, the predicted clean latent is decoded by the VAE decoder to produce the SR output $\hat{y}$. No loss is computed and no parameters are updated at test time.

The formulation above defines the base RareFlow model and its training objective. We next describe the RTS benchmark and the caption-construction pipeline used to instantiate the text-conditioning branch in practice.
\section{Data}
\label{sec:data}

\subsection{Benchmark Datasets}
We build a machine-learning-ready cross-sensor dataset for studying Arctic permafrost thaw as the primary benchmark in this work. Permafrost, defined as ground that remains below 0$^\circ$C for at least two consecutive summers, is undergoing rapid degradation due to climate change, leading to landscape instability and increased risks to critical infrastructure and communities \cite{liljedahl2016pan}. Monitoring these changes is therefore of significant scientific and societal importance.

A commonly studied indicator of permafrost thaw is RTS \cite{nitze2025darts, yang2023mapping}, with substantial variation in size (from a few tens to hundreds of meters), posing a challenge for detecting smaller, growing slumps in coarser-resolution public imagery such as Sentinel and Landsat. While VHR imagery (e.g., Maxar at >0.5 m resolution) can better capture these subtle features, it is prohibitively expensive to acquire at scale.

To address this challenge, we construct a cross-sensor dataset by pairing Sentinel-2 and Maxar imagery, building upon the RTS dataset introduced in \cite{yang2023mapping}. Specifically, Sentinel-2 Level-1C data corresponding to the coordinates of each Maxar image sample were selected and downloaded. For each Sentinel-2 sample, the image with the lowest cloud coverage between August 1 and August 31, 2020, was chosen. The Maxar images, originally in the NSIDC Sea Ice Polar Stereographic North coordinate system (EPSG:3413), were reprojected to WGS 84 (EPSG:4326) for processing in Google Earth Engine (GEE). This dataset serves as the main benchmark in our study, as it reflects an important real-world scientific application. Specifically, the task is to reconstruct HR observations from LR Sentinel-2 inputs under cross-sensor and cross-temporal conditions.

This cross-sensor benchmark covers seven Arctic regions: the Yamal and Gydan Peninsulas, the Lena River, and Kolguev Island in Russia, as well as Herschel Island, the Horton Delta, the Tuktoyaktuk Peninsula, and Banks Island in Canada. These sites span diverse land-cover conditions, including tundra, ice-rich permafrost bluffs, and coastal slopes, making the dataset well suited for evaluating fine-scale spatial reconstruction in the pan-Arctic environments. The RTS annotations were produced through careful manual digitization of Maxar imagery by domain experts, supported by multi-temporal cross-checks with Sentinel-2 and ArcticDEM.\cite{yang2023mapping}. 

\begin{figure}[h]
\centering
    \includegraphics[width=\linewidth]{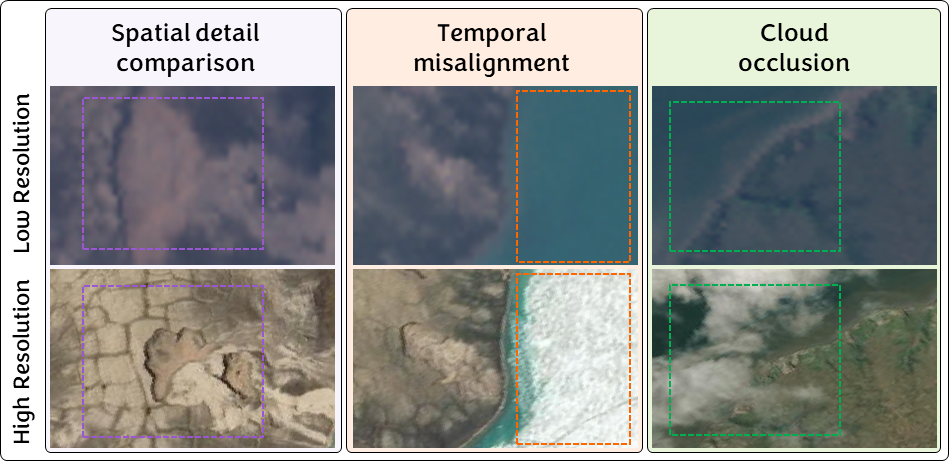}
    \caption{Examples of data challenges. From left to right, the columns illustrate: 1) spatial detail misalignment; 2) temporal misalignment (snow is present in HR but not LR); and 3) cloud occlusion in the HR image.}
    \label{fig:data_sample}
\end{figure}

This benchmark is particularly challenging because the LR-HR pairs do not align perfectly in time due to the limited revisit frequency of Maxar, which often provides only one cloud-free image per year, if not fewer, for a given area. As illustrated in Fig.~\ref{fig:data_sample}, the main challenges are fourfold. First, there is noticeable spatial and temporal misalignment between Sentinel-2 and Maxar acquisitions, which introduces sub-pixel shifts, illumination differences, seasonal variation, and, in some cases, real land-cover change, which can include the expansion of the RTS itself, as shown in Fig.~\ref{fig:data_sample} (a).  Second, the input images are often very small, with coverage of only a few hundred meters (e.g., 300 m by 400 m), corresponding to as few as $30 \times 40$ pixels in Sentinel-2 imagery, which poses challenges for general-purpose SR methods that typically target much larger image sizes. Third, Sentinel-2 imagery does not follow a standard 8-bit RGB radiometric range, which requires explicit normalization before it can be processed by the network. Fourth, the available training data are limited, with approximately 800 training images due to the the challenges in collecting labeled data. Together, these factors make RTS a highly out-of-distribution and data-constrained benchmark.

\begin{figure}[h]
  \centering
  \includegraphics[width=0.9\linewidth]{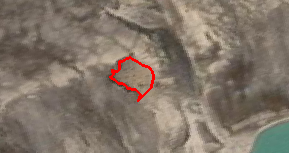}
  
  \vspace{1em}
  \begin{quote}
    \small
    \textbf{Generated Caption:} \itshape
    “A bowl-shaped thaw slump cuts into a rugged coastal hillside, exposing dark brown soil and crumbly earth, with pale sandy streaks sliding downslope toward blue-green water, surrounded by gray-brown, sparsely vegetated tundra.”
  \end{quote}
  
  \caption[VLM Caption Generation Example]{Representative output from the caption generation pipeline. The VLM (GPT-5) converts the visual features of the RTS into a natural language description, while avoiding technical jargon and spatial measurements.}
  \label{fig:data_re}
\end{figure}

\subsection{Caption Construction for Text Conditioning}
\label{subsec:caption_construction}

In addition to visual input, our framework incorporates textual guidance for RTS. We therefore generate captions directly from the HR reference images using a Vision-Language Model (VLM), GPT-5 \cite{openai2025gpt5}. The goal of these captions is to capture semantic and texture-level cues that may complement the limited spatial detail available in the LR Sentinel-2 input. To make this guidance useful for SR, the prompt is designed to emphasize visually observable structures, surface textures, and scene composition, while avoiding technical jargon, exact measurements, or unsupported statements about scale. In this way, the textual input remains descriptive enough to guide reconstruction, while staying grounded in the visible image content. In RareFlow, this text is not a replacement for the LR observation. It is the semantic prior that must be balanced against the LR-conditioned ControlNet pathway during reconstruction. 

It is important to note that these captions are used only during the training phase of the cross-sensor SR. During inference, only captions derived from the LR input, or a general caption, are provided to the model. This design avoids data leakage and ensures that captions serve as privileged semantic supervision during training, while test-time reconstruction relies on the LR observation and the learned RareFlow model, rather than on any target-derived description.

\begin{figure*}[t]
\begin{tcolorbox}[
    colback=black!3!white, 
    colframe=black!70!black, 
    fonttitle=\bfseries, 
    title=System Prompt Configuration for VLM,
    sharp corners=south, 
    boxrule=0.8pt,
    width=\textwidth
]
\small 
\textbf{Persona \& Objective}
\par
You are an expert satellite image analyst writing captions. Your task is to describe images of permafrost thaw slumps. The goal is to create natural, descriptive captions suitable for training a text-to-image AI model. The captions must sound like a human describing a photo in simple terms.

\medskip
\textbf{Style Rules}
\begin{itemize}[noitemsep, topsep=0pt]
    \item Use simple, everyday language.
    \item Write in a natural, fluid style.
    \item Use the present tense.
    \item Do not refer to "this image" or "the photo".
\end{itemize}

\medskip
\textbf{Content Requirements}
\begin{itemize}[noitemsep, topsep=0pt]
    \item \textbf{Main Feature:} Describe the thaw slump using common terms like "landslide," "thaw slump," "ground collapse," or "erosion scar."
    \item \textbf{Shape \& Form:} Mention its shape with simple descriptions like "crescent-shaped," "bowl-shaped," or "tongue of dirt."
    \item \textbf{Colors \& Textures:} Describe the colors and textures of the ground, vegetation, and water (\eg, "dark brown soil," "green tundra," "cracked earth," "blue-green ocean").
    \item \textbf{Setting:} Briefly describe the surrounding environment, such as "coastal cliff," "green hillside," "tundra plain," or "riverbank."
\end{itemize}

\medskip
\textbf{Exclusion Criteria (Crucial Constraints)}
\begin{itemize}[noitemsep, topsep=0pt]
    \item \textbf{NO JARGON:} Do not use technical terms like "RTS," "headwall," "lobe," or "rilled."
    \item \textbf{NO MEASUREMENTS:} Do not mention relative size, scale, or proportions (\eg, "covers a tenth of the scene," "a small feature").
    \item \textbf{NO ANNOTATIONS:} Do not mention map overlays, red lines, or other non-terrain elements.
\end{itemize}
\end{tcolorbox}
\caption{Detailed System Prompt provided to GPT-5.} 
\label{fig:system_prompt}
\end{figure*}

\begin{table*}[t]
\caption{Comparison of baseline methods, including their application domains, inputs, core ideas, and reasons for selection.}
\label{tab:baseline_comparison}
\centering
\small
\setlength{\tabcolsep}{3.5pt}
\renewcommand{\arraystretch}{1.15}
\begin{tabularx}{\textwidth}{p{1.7cm} p{1.35cm} p{1.6cm} p{4cm}  X X}
\toprule
\textbf{Model} & \textbf{Domain} & \textbf{Setting} & \textbf{Core idea} & \textbf{Input} & \textbf{Why useful as baseline}  
\\
\midrule

\textbf{SeeSR} \cite{wu2024seesr}
& General
& SISR
& Semantics-aware diffusion SR
& Single RGB image
& Tests whether semantic guidance improves perceptual SR quality
\\

\textbf{SAMSR} \cite{samsr}
& General
& SISR
& Semantic-guided diffusion with segmentation-aware constraints
& Single RGB image
& Strong semantic-guided diffusion baseline with explicit structural guidance
\\

\textbf{OpenSR} \cite{donike2025trustworthy}
& Remote sensing
& SISR
& Latent diffusion SR for Sentinel-2 with uncertainty-aware outputs
& Sentinel-2 RGB+NIR image
& Cross-Sensor remote-sensing diffusion baseline in a practical Earth-observation setting
\\

\textbf{MISR-S2} \cite{okabayashi2024cross}
& Remote sensing
& Multi-image SR
& Cross-sensor SR for irregular Sentinel-2 time series
& Irregular Sentinel-2 time series
& Closest task-level baseline: remote sensing, cross-sensor, and temporal
\\

\textbf{ZoomLDM} \cite{ZoomLDM}
& Generative modeling
& Multi-scale SR
& Multi-scale latent diffusion with scale-aware conditioning
& Multi-scale image patches
& Useful reference for multi-scale diffusion design and global structure modeling
\\

\textbf{AdcSR} \cite{chen2025adversarial}
& General
& Real-ISR
& Diffusion-GAN via adversarial distillation
& Single RGB image
& Strong baseline for testing whether compressed diffusion priors retain perceptual quality
\\

\bottomrule
\end{tabularx}
\end{table*}

\section{Experiments}
\label{sec:experiments}

We evaluate RareFlow through four questions: Does it improve the reconstruction quality under real cross-sensor mismatch? Does it benefit from the proposed semantic–observation (in the LR image) balance? Does it avoid test-time leakage from HR-derived text? And are its predictions reliable, rather than merely visually sharp?

\subsection{Experimental Setup}

\subsubsection{Compared Methods}
We compare RareFlow with two groups of baselines, summarized in Table~\ref{tab:baseline_comparison}. The first group consists of SR methods designed for remote sensing imagery and Sentinel-2-based inputs, including OpenSR \cite{donike2025trustworthy} and MISR-S2 \cite{okabayashi2024cross} (Although MISR-S2 was primarily developed for multi-image super-resolution, the authors also provide a pretrained checkpoint that enables single-image super-resolution (SISR) inference). The second group includes strong semantic-guided and generative SR models, namely AdcSR \cite{chen2025adversarial}, SAMSR \cite{samsr}, SeeSR \cite{wu2024seesr}, and ZoomLDM \cite{ZoomLDM}. Together, these baselines provide meaningful references for distortion-oriented, perceptual, and generative behavior under a difficult real-data setting.

\subsubsection{Implementation Details}
\paragraph{Training Configuration.}
To ensure stable convergence, we initialize the gating parameters such that the initial ControlNet influence is low, corresponding to an initial gate value of approximately $\alpha \approx 0.2$. We freeze the VAE, the SD3 Transformer backbone, and all text encoders, and optimize only the ControlNet parameters together with the gating module. Optimization is performed using AdamW. To reduce memory usage, we enable gradient checkpointing in the ControlNet blocks.

\paragraph{Data Processing and Augmentation.}
All inputs are resized to $512 \times 512$. During training, we apply paired random cropping and horizontal flipping to both the conditioning image and the target HR image. To avoid overfitting to the exact color statistics of the conditioning input, we additionally apply paired color jitter, affecting saturation and hue with probability $0.5$.

\paragraph{Text Conditioning.}
We use textual guidance for the benchmark dataset, where the region of interest is explicitly defined. Captions are generated from the HR reference imagery using GPT-5 \cite{openai2025gpt5}. Fig.~\ref{fig:system_prompt} shows the prompt given to GPT-5) and are used as semantic guidance during training.
At test time, RareFlow receives a generic prompt "high resolution, clear view.” This preserves the text-conditioning interface of the model without leaking HR-specific information at the test time.

\subsubsection{Inference Protocol and Reproducibility}
All qualitative and quantitative results are obtained using 30 denoising steps. For generative comparisons, we use a fixed inference configuration across methods to ensure fair evaluation. We release the checkpoints corresponding to the main RareFlow model, together with the principal ablation variants discussed in this section, including spectral-only and color-only alternatives where relevant.

Training requires approximately 12 hours on a single NVIDIA A100 (80\,GB) GPU with mixed precision (BF16/FP16). For stochastic evaluation, we generate repeated predictions under the same checkpoint and estimate predictive mean and variance from the sample set. These moments are then used for Gaussian NLL (Negative Log-Likelihood) and regression ECE (Expected Calibration Error). 

\subsection{Evaluation Protocol}
\subsubsection{Benchmark Setting}

The experiments in this section are conducted on the cross-sensor benchmark introduced in Section~\ref{sec:data}. The task is to reconstruct a 2\,m Maxar HR image from a 10\,m Sentinel-2 LR observation under real cross-sensor mismatch, temporal inconsistency, and limited training data. We adopt the fixed RGB setting, and all metrics are computed on the reconstructed HR images against the corresponding references.

\subsubsection{Evaluation Metrics}
Real-world cross-sensor SR cannot be evaluated reliably with a single metric. In this setting, the HR reference is not a perfectly controlled target: the LR and HR images come from different sensors, may exhibit temporal and spatial mismatch, and the HR image itself can contain blur, compression, or acquisition artifacts. As a result, a model can obtain favorable pixelwise scores simply by reproducing the smoothness of the reference, while still failing to recover perceptually meaningful or scientifically useful detail. For this reason, we evaluate performance using three complementary groups of metrics: signal fidelity, perceptual similarity,and no-reference realism. Table~\ref{tab:metrics-glance} summarizes the evaluation suite used throughout this section.

\begin{table*}
\centering
\caption{Summary of evaluation metrics. FR: full-reference. NR: no-reference. Set: dataset-level distributional metric. $\uparrow$ indicates higher is better; $\downarrow$ indicates lower is better.}
\label{tab:metrics-glance}
\begin{tabularx}{0.93\linewidth}{l c c c X}
\toprule
Metric & Type & Range & Trend & Main role \\
\midrule
PSNR & FR & $[0,\infty)$ dB & $\uparrow$ & Pixel-level fidelity based on mean squared error; favors conservative reconstructions. \\
SSIM & FR & $[0,1]$ & $\uparrow$ & Structural similarity based on luminance, contrast, and structure. \\
SAM & FR & $[0^\circ,180^\circ]$ & $\downarrow$ & Angular difference between spectral vectors to measure spectral consistency. \\
LPIPS & FR & $\ge 0$ & $\downarrow$ & Perceptual distance in deep feature space. \\
DISTS & FR & $\ge 0$ & $\downarrow$ & Joint structural and textural similarity. \\
FID & Set & $\ge 0$ & $\downarrow$ & Distributional distance between generated and reference sets. \\
NIQE & NR & $\ge 0$ & $\downarrow$ & Deviation from natural-scene statistics. \\
MANIQA & NR & model dep. & $\uparrow$ & Transformer-based no-reference quality assessment. \\
\bottomrule
\end{tabularx}
\end{table*}

We first report full-reference fidelity metrics. PSNR (Peak Signal-to-Noise Ratio) and SSIM (Structural Similarity Index Measure) measure pixelwise reconstruction accuracy and local structural agreement. We also report SAM (Spectral Angle Mapper), which measures angular disagreement between the reconstructed and reference color vectors and is therefore useful for assessing cross-sensor spectral consistency. 

To address this limitation, we include deep perceptual metrics such as LPIPS (Learned Perceptual Image Patch Similarity) and DISTS (Deep Image Structure and Texture Similarity). These metrics are better aligned with human visual judgment and are more sensitive to texture quality, structural sharpness, and perceptual plausibility than pure pixelwise distortion.

A challenge with all full-reference metrics is their reliance on a perfect HR ground truth, which is rarely available in real remote sensing scenarios. This motivates the use of no-reference, or blind, image quality metrics. NIQE (Natural Image Quality Evaluator) \cite{niqe} measures how far an image deviates from the statistics of natural scenes, providing a reference-free estimate of realism. While NIQE offers a general signal, modern blind quality models can capture perceptual quality more accurately. We therefore also report MANIQA (Multi-dimension Attention Network for Image Quality Assessment) \cite{maniqa}, a Transformer-based image quality assessor that is designed to align more closely with human opinion.

Because RareFlow is a generative model, image-level quality alone is still not enough. A strong generative SR model should also match the overall distribution of real HR images rather than only optimize individual samples. For this reason, we report Fr\'echet Inception Distance (FID) \cite{fid}, which compares the feature distributions of generated and reference image sets. FID therefore complements LPIPS and NIQE by measuring whether the model produces a realistic and coherent output distribution at the dataset level.

For an HR reference image $y \in \mathbb{R}^{H\times W}$ and reconstruction $\hat{y} \in \mathbb{R}^{H\times W}$, PSNR is
\begin{equation}
\mathrm{PSNR}(y,\hat{y})=10\log_{10}\left(\frac{R^2}{\frac{1}{HW}\sum_{i,j}(y_{ij}-\hat{y}_{ij})^2}\right),
\end{equation}
where $R$ is the peak intensity value, and $H$ and $W$ denote image height and width. We compute PSNR on the luminance channel.

SSIM \cite{ssim} measures local structural agreement:
\begin{equation}
\mathrm{SSIM}(y,\hat{y})=
\frac{(2\mu_y\mu_{\hat{y}}+C_1)(2\sigma_{y\hat{y}}+C_2)}
{(\mu_y^2+\mu_{\hat{y}}^2+C_1)(\sigma_y^2+\sigma_{\hat{y}}^2+C_2)}.
\end{equation}
Where $\mu_y$ and $\mu_{\hat y}$ are local means, $\sigma_y^2$ and $\sigma_{\hat y}^2$ are local variances, $\sigma_{y\hat y}$ is the local covariance, and $C_1,C_2$ are stabilization constants. Image-level SSIM is obtained by averaging over spatial windows.

For an RGB or spectral vector at pixel $p$, SAM is computed as
\begin{equation}
\mathrm{SAM}(y,\hat{y}) =
\frac{1}{N}\sum_{p=1}^{N}
\arccos
\left(
\frac{y_p^\top \hat{y}_p}
{\|y_p\|_2 \|\hat{y}_p\|_2 + \epsilon}
\right),
\end{equation}
where $N$ is the number of pixels and $\epsilon$ is a small constant for numerical stability.

LPIPS \cite{lpips} measures perceptual distance in deep feature space:
\begin{equation}
d_{\mathrm{LPIPS}}(y,\hat{y})=\sum_{l}\frac{1}{H_lW_l}\sum_{h,w}
\left\|w_l \odot \big(\hat\phi_l^{h,w}(y)-\hat\phi_l^{h,w}(\hat{y})\big)\right\|_2^2.
\end{equation}
Here, $l$ indexes feature layers, $H_l$ and $W_l$ are the spatial dimensions of layer $l$, $\hat{\phi}_l^{h,w}(\cdot)$ denotes the channel-normalized deep feature vector at location $(h,w)$, $w_l$ are learned channel weights, and $\odot$ denotes elementwise multiplication.

FID \cite{fid} evaluates the distributional gap between sets of reconstructed and reference images:
\begin{equation}
\mathrm{FID}
=
\|\mu_y-\mu_{\hat{y}}\|_2^2
+
\mathrm{Tr}\!\Big(
\Sigma_y+\Sigma_{\hat{y}}-2(\Sigma_y\Sigma_{\hat{y}})^{1/2}
\Big),
\end{equation}
where $(\mu_y,\Sigma_y)$ and $(\mu_{\hat{y}},\Sigma_{\hat{y}})$ are the empirical mean and covariance of Inception features computed from the reference and reconstructed image sets, respectively, $\mathrm{Tr}(\cdot)$ denotes the trace, and $(\cdot)^{1/2}$ denotes the matrix square root.
\subsubsection{Human Expert Study Protocol}

In addition to numerical benchmarks, we conduct a domain-expert study to assess whether the generated super-resolution images are scientifically meaningful in reconstructing rare Earth features (i.e., RTS in this benchmark dataset). The evaluation is performed by researchers working in Arctic permafrost science and remote sensing, including Ph.D.-level research scientists and senior research assistants experienced in identifying thaw slump features from high-resolution imagery.

\begin{figure*}
    \centering
    \includegraphics[width=1\linewidth]{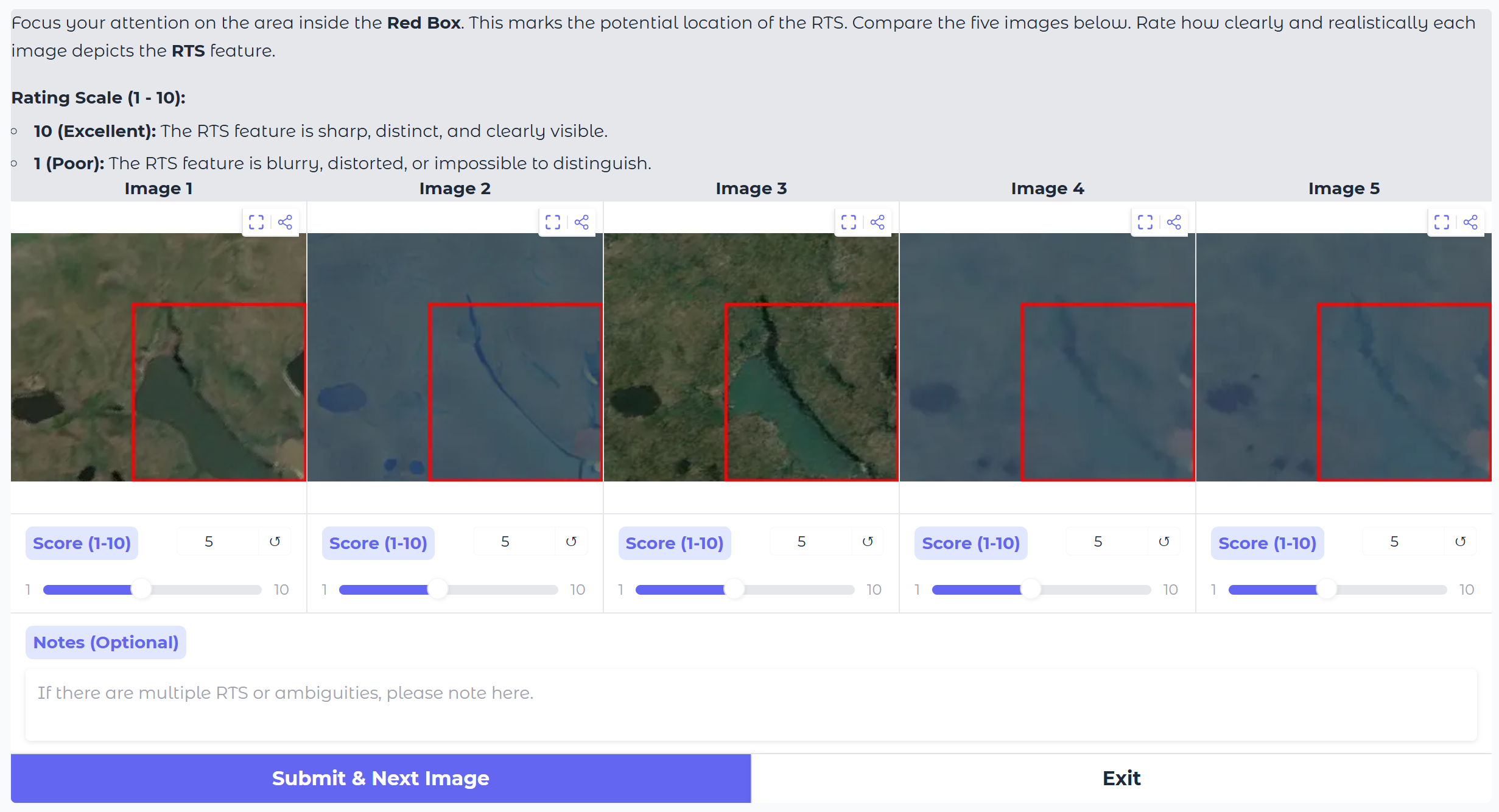}
    \caption{Web interface used for expert evaluation.}
    \label{fig:human_eval_interface}
\end{figure*}

Each expert evaluates 30 randomly sampled RTS scenes using a custom web interface, shown in Fig.~\ref{fig:human_eval_interface}. For each scene, the interface displays five aligned but anonymized images of the same area: the HR reference, the original Sentinel-2 LR input, the RareFlow reconstruction, and the outputs of the two best-performing baseline methods. A red box marks the candidate RTS region, and participants are instructed to focus their assessment on that area. They then rate each image independently on a 1--10 scale according to how clearly and realistically it depicts the RTS feature, with 10 indicating clearly visible feature, and 1 indicating a blurry or indistinguishable feature.

\subsection{Main Results}

Table~\ref{tab:sr_models_trimmed} compares RareFlow with several baseline models for cross-sensor SR using fidelity, perceptual, and no-reference metrics. RareFlow achieves the best result on six of the eight metrics, including SAM (3.86), LPIPS (0.36), DISTS (0.30), and FID (116.16). Compared with the next-best SAM score among the baselines, SeeSR at 12.26, RareFlow reduces SAM by about 68.52\% indicating its generated pixels are more closely aligned with the spectral vectors of the Maxar target, which helps preserve material signatures across the sensor gap. RareFlow also improves FID relative to the next-best method, AdcSR at 187.18, with a reduction of approximately ~38\%. This suggests that the overall image statistics of RareFlow outputs are much closer to the Maxar target distribution.

At the same time, RareFlow remains competitive on conventional fidelity metrics. It achieves the best SSIM (0.59), while its PSNR (18.76\,dB) is effectively tied with the strongest baseline result (18.78\,dB for SeeSR). This pattern is consistent with the perception-distortion trade-off \cite{blau2018perception}, where methods optimized primarily for pixelwise reconstruction may score slightly higher on PSNR while producing visibly smoother outputs. The strength of RareFlow is that it improves perceptual realism without materially sacrificing structural faithfulness (as reflected in its strong PSNR).

\begin{figure*}[t]
    \centering
    \includegraphics[width=1\linewidth]{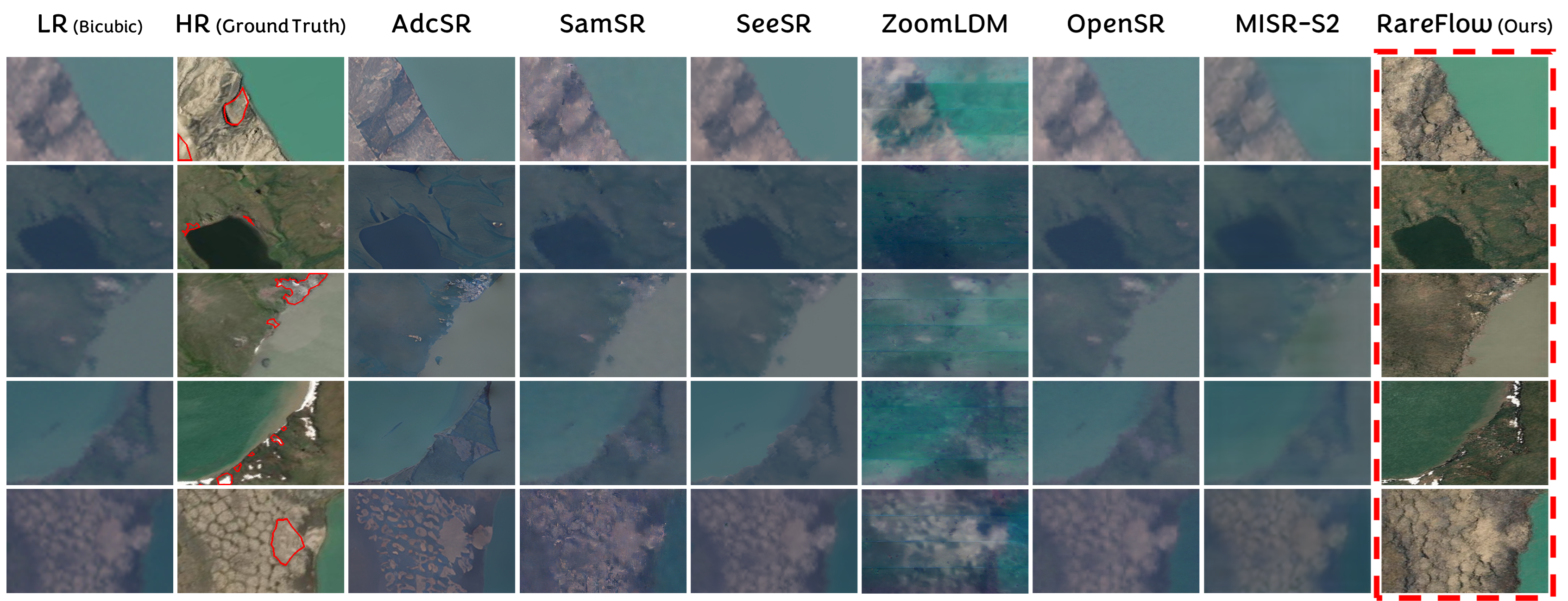}
    \caption{Qualitative comparison on paired LR–HR data. Features annotated with a red outline are retrogressive thaw slumps.}
    \label{fig:main_results}
\end{figure*}

Beyond pure resolution enhancement, the core challenge of this task is performing simultaneous SR and cross-sensor style transfer. The model must also bridge the appearance gap between Sentinel-2 and Maxar imagery. Fig.~\ref{fig:main_results} illustrates this point qualitatively. Several baselines preserve the low-frequency content of the LR image but retain the muted contrast and smoother texture profile characteristic of Sentinel-2. In contrast, RareFlow reconstructs sharper geomorphological detail while also translating the image toward the target Maxar appearance, yielding outputs that are both more informative and more visually consistent with the HR reference.

\begin{table*}[h!]
\centering
\footnotesize
\setlength{\tabcolsep}{4pt}
\caption{SR results on \textbf{paired LR–HR} data. \textbf{Bold} = best, \underline{underline} = second-best; arrows indicate whether higher or lower is better. Performance Gain (\%) row reports the relative change of RareFlow with respect to the best baseline for each metric.}
\label{tab:sr_models_trimmed}

\sisetup{detect-weight, mode=text, round-mode=places, round-precision=2}

\newcommand{\gain}[1]{\textcolor{green!50!black}{#1}}
\newcommand{\drop}[1]{\textcolor{red!70!black}{#1}}

\begin{tabular}{l|
                S[table-format=2.2]
                S[table-format=1.2]
                S[table-format=2.2] |
                S[table-format=1.2]
                S[table-format=1.2] |
                S[table-format=3.2]
                S[table-format=2.2]
                S[table-format=1.2]
                }
\toprule
\multicolumn{1}{c|}{\multirow{2}{*}{\textbf{SR Model}}} & \multicolumn{3}{c|}{\textbf{Fidelity Metrics}} & \multicolumn{2}{c|}{\textbf{Perceptual Similarity}} & \multicolumn{3}{c}{\textbf{Realism Metrics}} \\
\cmidrule(lr){2-4} \cmidrule(lr){5-6} \cmidrule(lr){7-9}
& {PSNR $\uparrow$} & {SSIM $\uparrow$} & {SAM $\downarrow$} & {LPIPS $\downarrow$} & {DISTS $\downarrow$} & {FID $\downarrow$} & {NIQE $\downarrow$} & {MANIQA $\uparrow$} \\
\midrule
Bicubic         & 18.22  & 0.54 & 12.58 & 0.71 & 0.40 & 205.35 & 12.42 & 0.27 \\
ZoomLDM         & 17.23 & 0.26 & 12.96 & 0.60 & 0.59 & 352.11 & 18.10 & 0.19 \\
SeeSR           & \textbf{18.78} & 0.50 & {\underline{12.26}} & 0.46 & 0.38 & 302.36 & 10.78 & \textbf{0.36} \\
AdcSR           & 18.59 & \underline{0.58} &  12.31 & \underline{0.40} & 0.37 & \underline{187.18} & \underline{8.38} & 0.28 \\
MISR-S2         & 18.39 & 0.50 & 12.72 & 0.54 & 0.43 & 254.70 & 13.55 & \underline{0.33} \\
SAMSR           & 18.36 & 0.54 & 12.80 & 0.48 & 0.39 & 189.01 & 11.84 & 0.32 \\
OpenSR          & 17.29 & 0.51 & 12.59 & 0.41 & \underline{0.36} & 225.62 & 9.80 & 0.25 \\
RareFlow (Ours) & \underline{18.76} & \textbf{0.59} & \textbf{3.86} & \textbf{0.36} & \textbf{0.30} & \textbf{116.16} & \textbf{5.36} & 0.31 \\
\midrule
\multicolumn{1}{l|}{\textbf{Performance Gain (\%)}} &
\multicolumn{1}{c}{\drop{$-0.11$}} &
\multicolumn{1}{c}{\gain{$1.72$}} &
\multicolumn{1}{c|}{\gain{$68.52$}} &
\multicolumn{1}{c}{\gain{$10.00$}} &
\multicolumn{1}{c|}{\gain{$16.67$}} &
\multicolumn{1}{c}{\gain{$37.94$}} &
\multicolumn{1}{c}{\gain{$36.04$}} &
\multicolumn{1}{c}{\drop{$-13.89$}} \\
\bottomrule
\end{tabular}
\end{table*}

\begin{table*}[t]
\centering
\footnotesize
\setlength{\tabcolsep}{2pt}
\caption{
Ablation study of RareFlow components using fidelity, perceptual-similarity, and realism metrics. 
\textbf{Bold} = best, \underline{underline} = second-best; arrows indicate whether higher or lower is better.
}
\label{tab:ablation}
\sisetup{detect-weight, mode=text, round-mode=places, round-precision=2}
\begin{tabular}{l|
                S[table-format=2.2]
                S[table-format=1.2]
                S[table-format=1.2] |
                S[table-format=1.2]
                S[table-format=1.2] |
                S[table-format=3.2]
                S[table-format=1.2]
                S[table-format=1.2]
                }
\toprule
\multicolumn{1}{c|}{\multirow{2}{*}{\textbf{Model Configuration}}} 
& \multicolumn{3}{c|}{\textbf{Fidelity Metrics}} 
& \multicolumn{2}{c|}{\textbf{Perceptual Similarity}} 
& \multicolumn{3}{c}{\textbf{Realism Metrics}} \\
\cmidrule(lr){2-4} \cmidrule(lr){5-6} \cmidrule(lr){7-9}
& {PSNR $\uparrow$} 
& {SSIM $\uparrow$} 
& {SAM $\downarrow$} 
& {LPIPS $\downarrow$} 
& {DISTS $\downarrow$} 
& {FID $\downarrow$} 
& {NIQE $\downarrow$} 
& {MANIQA $\uparrow$} \\
\midrule
(1) Baseline (ControlNet from scratch) 
& 18.01 & 0.49 & 6.54 & 0.41 & 0.33 & 206.32 & 5.98 & 0.25 \\

(2) Pre-trained ControlNet 
& \underline{18.80} & 0.51 & 6.10 & \textbf{0.35} & \textbf{0.29} & 187.60 & 6.42 & 0.19 \\

(3) Pre-trained ControlNet + Caption 
& 17.08 & 0.38 & 4.88 & 0.47 & 0.32 & 145.43 & 5.64 & \textbf{0.32} \\

(4) Pre-trained ControlNet + Caption + $\alpha$-gate 
& 17.69 & \underline{0.52} & 4.53 & 0.38 & \underline{0.30} & \underline{138.21} & 5.78 & 0.23 \\

(5) Pre-trained ControlNet + Loss 
& \textbf{19.11} & 0.50 & 5.48 & 0.41 & 0.33 & 159.47 & 6.00 & 0.22 \\

(6) Pre-trained ControlNet + Caption + Loss 
& 17.11 & 0.38 & \textbf{3.48} & 0.47 & 0.32 & 144.61 & \underline{5.61} & \textbf{0.32} \\

\midrule
\textbf{(7) Full Model (Pre-trained ControlNet + Caption + Loss + $\alpha$-gate)} 
& 18.76 & \textbf{0.59} & \underline{3.86} & \underline{0.36} & \underline{0.30} & \textbf{116.16} & \textbf{5.36} & \underline{0.31} \\
\bottomrule
\end{tabular}
\end{table*}

\begin{figure*}
    \centering
    \includegraphics[width=1\linewidth]{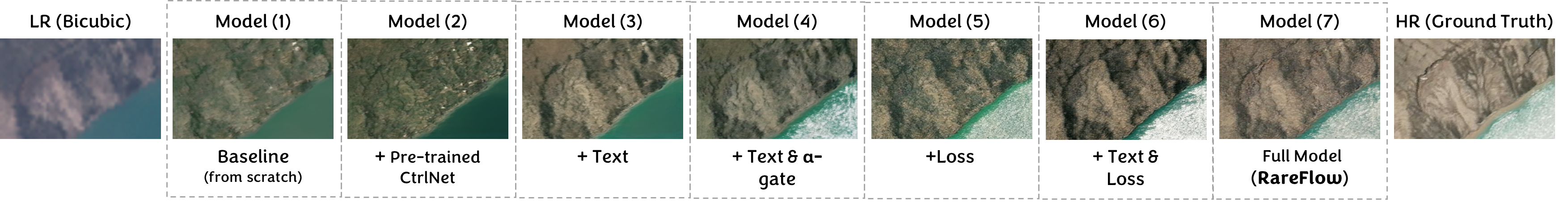}
    \caption{Visual Comparison of model variants.}
    \label{fig:ablation}
\end{figure*}

\subsection{Ablation Studies}

We study the contribution of three components in RareFlow: (i) structural conditioning through ControlNet, including pre-trained initialization and $\alpha$-gated conditioning, (ii) semantic guidance from text captions, and (iii) the consistency-guided objective. Table~\ref{tab:ablation} reports their effect on fidelity, perceptual similarity, and realism metrics.

The results show a clear trade-off between fidelity-oriented metrics and realism-oriented metrics. Variants with stronger structural conditioning improves fidelity and perceptual-similarity metrics, indicating that structural initialization helps preserve the paired image content, while variants with caption guidance improve realism and spectral accuracy, as reflected by lower FID, lower SAM, and higher MANIQA, but it also reduces PSNR and SSIM. 
This indicates that semantic guidance helps move the output toward the Maxar appearance, but can weaken pixel-level agreement with the paired reference. The consistency-guided objective does not act as a direct perceptual enhancement module. 
Instead, it mainly stabilizes the reconstruction and helps constrain the caption-guided generation. The full model, which combines all components, provides the best balance. Fig. \ref{fig:ablation} shows a visual comparison of super-resolved results from models under different ablation conditions.

\subsubsection{Effect of ControlNet Initialization and Gating}

The effect of ControlNet is evident from the improvement of variant (2) over the baseline model (1).
In Table~\ref{tab:ablation}, pre-training improves PSNR from $18.01$ to $18.80$, SSIM from $0.49$ to $0.51$, SAM from $6.54$ to $6.10$, LPIPS from $0.41$ to $0.35$, and DISTS from $0.33$ to $0.29$. These improvements confirm that pre-trained structural conditioning provides a stronger starting point than learning the conditioning from scratch.

However, fixed structural conditioning can also be restrictive. When the LR input contains blur or local softness, a strong conditioning signal may encourage the model to preserve these degraded details. The proposed $\alpha$-gate addresses this issue by adaptively controlling the strength of the conditioning signal. This allows the model to preserve scene structure while reducing over-dependence on low-quality local details.

The effect of gating is visible when comparing caption-guided variants with and without the gate. Adding the $\alpha$-gate to the caption-guided model improves PSNR, SSIM, SAM, LPIPS, DISTS, and FID, showing that adaptive conditioning improves both fidelity and realism in this setting.

A similar pattern appears when the gate is added on top of both caption guidance and the consistency-guided objective. 
Compared with variant (6), the full model improves PSNR, SSIM, LPIPS, DISTS, FID, and NIQE. 
At the same time, SAM and MANIQA show a small trade-off. 
This confirms that $\alpha$-gating helps balance structural preservation and realistic generation.

\subsubsection{Effect of Semantic Guidance}
 Caption guidance mainly improves realism and spectral accuracy. In Table~\ref{tab:ablation}, adding captions to the pre-trained ControlNet model reduces FID from $187.60$ in variant (2) to $145.43$ in variant (3), and increases MANIQA from $0.19$ to $0.32$. SAM also improves from $6.10$ to $4.88$. 
 At the same time, PSNR decreases from $18.80$ to $17.08$, and SSIM decreases from $0.51$ to $0.38$. 

 This improvement comes with a clear fidelity cost. 
PSNR decreases from $18.80$ to $17.08$, and SSIM decreases from $0.51$ to $0.38$. 
 This result indicates that semantic guidance is beneficial for recovering plausible target-domain appearance detail, but that it can also encourage the model to generate content that is semantically reasonable while being less well aligned with the underlying image evidence.

\subsubsection{Effect of Consistency-Guided Loss}
The consistency-guided objective has a different role from caption guidance. 
Without captions, comparing variant (2) with variant (5) shows that the loss improves PSNR from $18.80$ to $19.11$, SAM from $6.10$ to $5.48$, and FID from $187.60$ to $159.47$. 
These changes indicate better agreement with the paired target and closer overall image statistics. 
However, LPIPS, DISTS, NIQE, and MANIQA do not improve. 
This suggests that the loss alone stabilizes reconstruction, but is not sufficient to produce stronger perceptual realism.

The role of the consistency-guided objective becomes clearer when captions are used. 
Caption guidance alone improves realism metrics but reduces pixel-level fidelity. 
When the consistency-guided objective is added to the caption-guided model, SAM improves from $4.88$ to $3.48$. 
This suggests that the loss helps constrain caption-guided generation toward the paired target while preserving most of the realism gain introduced by captions.

The full model combines semantic guidance, consistency-guided training, and $\alpha$-gated structural conditioning. 
This combination gives the strongest overall result: the model achieves the best SSIM, FID, and NIQE, and remains second-best or tied second-best on SAM, LPIPS, DISTS, and MANIQA. 
These results suggest that the components are complementary. 
Pre-trained ControlNet supports structural consistency, captions improve target-domain realism, the consistency-guided objective stabilizes caption-guided generation, and the $\alpha$-gate helps balance conditioning strength.

\subsection{Effect of Different Prompt Choices for Semantic-guided Generation}
During the training phase, although we have access to the HR ground truth image, semantic guidance cannot be obtained by directly prompting the HR image. To avoid data leakage, we need to adopt alternative prompting methods for semantic guidance. Hence, we evaluated three test-time prompting strategies that do not rely on any information derived from HR images: (i) a general text guidance (\emph{"high resolution, clear view"}), (ii) no text, and (iii) a caption generated from the LR image alone. As shown in Table~\ref{tab:sem}, the model remains competitive without any HR-derived text at test time. Among these settings, the general prompt (i) gives the best overall performance. Therefore, we use the general prompt, “high resolution, clear view,” for all reported experiments.

\begin{table*}[h]
    \centering
    \footnotesize
    \caption{Test-time prompt ablation for the full model under three semantic-guidance settings.}
    {
    \begin{tabular}{l|ccc|cc|ccc}
        \toprule
        \multicolumn{1}{c|}{\multirow{2}{*}{\textbf{Semantic Guidance}}} 
        & \multicolumn{3}{c|}{\textbf{Fidelity Metrics}} 
        & \multicolumn{2}{c|}{\textbf{Perceptual Similarity}} 
        & \multicolumn{3}{c}{\textbf{Realism Metrics}} \\
        \cmidrule(lr){2-4} \cmidrule(lr){5-6} \cmidrule(lr){7-9} 
        & {PSNR $\uparrow$} & {SSIM $\uparrow$} & {SAM $\downarrow$} 
        & {LPIPS $\downarrow$} & {DISTS $\downarrow$} 
        & {FID $\downarrow$} & {NIQE $\downarrow$} & {MANIQA $\uparrow$} \\
        \midrule
        General text
        & 18.76 & 0.59 & 3.86 & 0.36 & 0.30 & 116.16 & 5.36 & 0.31\\
        No-text
        & 18.28 & 0.38 & 3.91 & 0.45 & 0.34 & 129.32 & 5.58 & 0.31\\
        LR-only
        & 17.46 & 0.37 & 5.28 & 0.48 & 0.32 & 168.29 & 6.05 & 0.27\\
        \bottomrule
    \end{tabular}
    }
    \label{tab:sem}
\end{table*}

\subsection{Expert Evaluation}

\begin{figure}[t]
    \centering
    \includegraphics[width=.95\linewidth]{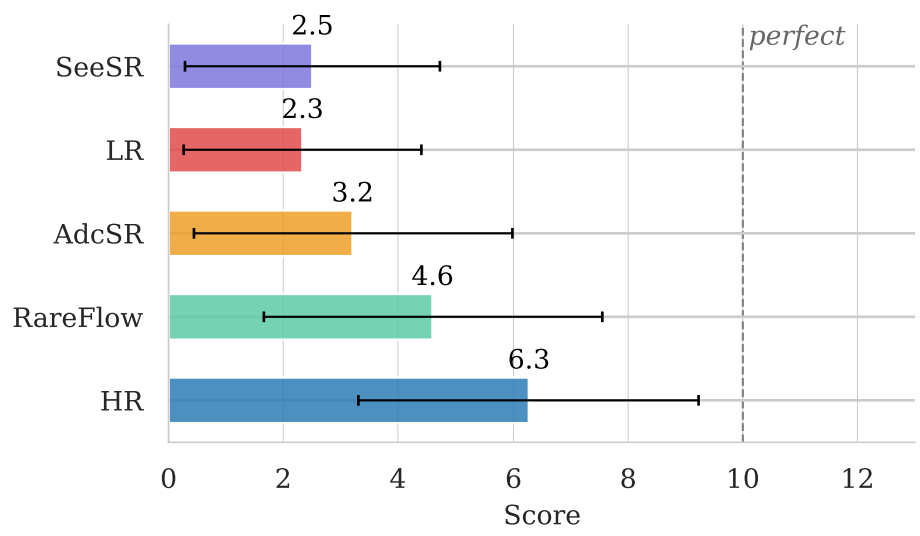}
    \caption{
Expert evaluation on the quality of the reconstructed images. The bar chart shows the mean perceptual score ± standard deviation for each method (RareFlow, AdcSR, SeeSR), as well as the low-resolution (LR) and high-resolution (HR) images, based on a 1–10 scale (the dashed line marks the perfect score of 10).
        }
    \label{fig:expert_eval}
\end{figure}

To complement the quantitative benchmarks, we conduct an expert evaluation on the RTS benchmark. A total of five experts, including Earth scientists and data scientists working on RTS, were invited to compare RareFlow with the LR input, the HR reference, and the two best-performing baseline SR methods, AdcSR and SeeSR. Each image was scored from 1 to 10 based on overall perceptual quality, with attention to RTS boundaries.

The mean perceptual scores, along with their standard deviations, are illustrated in Fig. \ref{fig:expert_eval}. RareFlow achieved a mean score of 4.6, outperforming the competing methods AdcSR (3.2) and SeeSR (2.5), and substantially narrowing the gap between the LR baseline (2.3) and the HR ground truth (6.3).

\begin{table}[t]
    \centering
    \caption{Inter-annotator agreement for the expert perceptual evaluation. All metrics range between 0 and 1. Higher values indicate stronger agreement.}
    \label{tab:expert_agreement}
    \scriptsize
    \begin{tabular}{@{}l c l@{}}
        \toprule
        \textbf{Metric} & \textbf{Value} & \textbf{Interpretation} \\
        \midrule
        Krippendorff's $\alpha$ & 0.736 & Substantial agreement \\
        Kendall's $W$ & 0.841 & Strong consistency \\
        Spearman's $\rho$ & 0.673 & Moderate-to-strong correlation \\
        \bottomrule
    \end{tabular}
\end{table}

To assess the reliability of the subjective scores, we report inter-annotator agreement in Table~\ref{tab:expert_agreement}. 
We use Krippendorff's $\alpha$ \cite{hayes2007krippendorff} to measure overall agreement while accounting for chance, Kendall's $W$ \cite{kendall1939rankings} to measure consistency in image rankings, and Spearman's $\rho$ \cite{spearman1904association} to measure pairwise correlation between annotators' scores. 
The resulting values indicate substantial agreement among the experts, suggesting that the scores are reasonably consistent across raters.

Experts judged RareFlow outputs to provide a clear improvement over the LR inputs. In many cases, RareFlow reconstructions were considered closer to the 2\,m Maxar reference than the competing SR methods, especially in preserving thaw-slump boundaries, exposed soil regions, and local terrain texture. However, HR remains the upper reference and still receives the highest overall scores.

The first stage assesses the validity of the semantic guidance used for text conditioning. For each sample, GPT-5~\cite{openai2025gpt5} is given the HR reference image and the RTS region of interest, and is asked to produce a caption describing surface texture, shape, and the surrounding local context.

We additionally assess the validity of the semantic guidance used for text conditioning. Experts evaluate whether the generated captions are grounded in visible geomorphological features rather than vague or unsupported descriptions. Each caption is assigned to one of four categories: perfect, minor error, major error, or irrelevant. The review showed that 57\% of the captions were rated as perfect, accurately describing the visible RTS morphology and providing useful guidance for reconstruction, while the remaining 43\% were rated as containing minor errors. No captions were classified as having major errors or as irrelevant. The minor inaccuracies may partly reflect the rarity of RTS landforms and their likely underrepresentation in the visual data used to train general-purpose vision-language and generative models. Consequently, while many captions captured the main slump structure well, some missed fine-scale features such as subtle boundary variations, exposed-soil texture, or nearby terrain details.

\section{Model Generalizability}
\label{sec:generalization}

To evaluate whether the proposed method transfers beyond the current benchmark data for environmental analysis, we trained and tested it on two public cross-sensor remote sensing super-resolution benchmarks: BreizhSR \cite{okabayashi2024cross} and SEN2NAIP \cite{aybar2024sen2naip}. Both benchmarks follow the same overall task formulation, namely super-resolution from Sentinel-2 LR input to a higher-resolution RGB target, but they differ substantially in scene content, geographic coverage, and sensor pairing. Together, they provide a useful test of generalization under distinct real-world image formation conditions.

 Unlike our benchmark data, these datasets do not define explicit feature of interest within the image scenes. We therefore do not generate dataset-specific VLM captions for them. Instead, we retain the RareFlow text-conditioning interface through a fixed generic prompt and let LR-conditioned observation control carry the main dataset-specific information.
 
\begin{table*}[t]
    \centering
    \caption{Results on public generalization benchmarks. \textbf{Bold} indicates the best result.}
    \setlength{\tabcolsep}{5pt}
    \scalebox{1}{
    \begin{tabular}{l|l|ccc|cc|ccc}
        \toprule
        \multirow{2}{*}{\textbf{Dataset}} 
        & \multirow{2}{*}{\textbf{Method}} 
        & \multicolumn{3}{c|}{\textbf{Fidelity Metrics}} 
        & \multicolumn{2}{c|}{\textbf{Perceptual Similarity}} 
        & \multicolumn{3}{c}{\textbf{Realism Metrics}} \\
        \cmidrule(lr){3-5} \cmidrule(lr){6-7} \cmidrule(lr){8-10}
        & & {PSNR $\uparrow$} & {SSIM $\uparrow$} & {FSIM $\uparrow$} 
        & {LPIPS $\downarrow$} & {DISTS $\downarrow$} 
        & {FID $\downarrow$} & {NIQE $\downarrow$} & {MANIQA $\uparrow$} \\
        
        \midrule
        \multirow{2}{*}{\centering BreizhSR}
        & RareFlow & \textbf{11.30} & 0.18 & 0.53 & \textbf{0.52} & \textbf{0.31} & \textbf{245.30} & \textbf{7.95} & \textbf{0.32} \\
        & MISR-S2  & 11.28 & \textbf{0.23} & \textbf{0.54} & 0.63 & 0.36 & 254.10 & 9.36 & 0.22 \\

        \midrule
        \multirow{2}{*}{\centering SEN2NAIP}
        & RareFlow & \textbf{14.44} & \textbf{0.29} & \textbf{0.75} & \textbf{0.59} & \textbf{0.32} & \textbf{214.50} & \textbf{6.23} & 0.19 \\
        & OpenSR   & 12.50 & 0.24 & 0.61 & 0.61 & 0.36 & 230.95 & 6.76 & \textbf{0.20} \\
        \bottomrule
    \end{tabular}
    }
    \label{tab:generalization_metrics}
\end{table*}

\subsection{BreizhSR}

BreizhSR \cite{okabayashi2024cross} is a cross-sensor remote sensing SR benchmark built over Brittany, France. It targets 4$\times$ super-resolution by pairing Sentinel-2 image time series with higher-resolution SPOT-6/7 observations. The dataset is dominated by agricultural and coastal landscapes, where performance depends on recovering thin parcel boundaries, shoreline geometry, and repeated texture patterns.

The LR source consists of cloud-free Sentinel-2 Level-2A RGB imagery at 10\,m resolution, while the HR reference is obtained from pansharpened SPOT-6/7 color imagery resampled to 2.5\,m \cite{okabayashi2024cross}. Although the acquisition window is constrained to reduce seasonal inconsistency, the benchmark still reflects a realistic cross-sensor setting with differences in geometry, radiometric response, and temporal sampling. In addition, the time-series design leads to irregular image availability across locations and dates.

For evaluation, we train RareFlow directly on BreizhSR and compare it with MISR-S2 on the same benchmark. This provides a matched comparison against a method developed specifically for Sentinel-2-to-SPOT super-resolution.

As shown in Table~\ref{tab:generalization_metrics}, RareFlow achieves the best FID, PSNR, LPIPS, DISTS, and MANIQA on BreizhSR, while MISR-S2 performs better on SSIM and FSIM. In particular, RareFlow improves FID from 254.10 to 245.30 and slightly increases PSNR from 11.28\,dB to 11.30\,dB. It also reduces LPIPS from 0.63 to 0.52 and DISTS from 0.36 to 0.31, while improving MANIQA from 0.22 to 0.32. Overall, these results indicate that RareFlow produces reconstructions with stronger perceptual quality and better distributional alignment to the HR target, even relative to a benchmark-specific baseline.

\subsection{SEN2NAIP}

SEN2NAIP \cite{aybar2024sen2naip} is a large-scale cross-sensor Sentinel-2 super-resolution benchmark constructed from correspondences between Sentinel-2 imagery and National Agriculture Imagery Program (NAIP) orthophotos. It provides broad geographic coverage across the contiguous United States and includes a wide range of land-cover types, including agricultural, suburban, and mixed rural scenes.

We use the cross-sensor portion of SEN2NAIP, where the LR input comes from Sentinel-2 at 10\,m resolution and the HR reference comes from NAIP imagery prepared at 2.5\,m resolution, yielding a 4$\times$ super-resolution task \cite{aybar2024sen2naip}. Despite careful curation to keep the temporal gap small, the benchmark remains challenging because the paired images differ in viewing geometry, sensor characteristics, texture statistics, and local registration quality. As a result, it represents a realistic cross-sensor reconstruction problem rather than a synthetic degradation setting.

For evaluation, we train RareFlow directly on SEN2NAIP and compare it with OpenSR on the same benchmark to provide a matched comparison. The results in Table~\ref{tab:generalization_metrics} show that RareFlow outperforms OpenSR on nearly all reported metrics. Specifically, RareFlow improves FID from 230.95 to 214.50, increases PSNR from 12.50\,dB to 14.44\,dB, and raises SSIM from 0.24 to 0.29. It also reduces LPIPS from 0.61 to 0.59 and DISTS from 0.36 to 0.32, while improving FSIM from 0.61 to 0.75. OpenSR obtains a slightly higher MANIQA score (0.20 versus 0.19), but overall RareFlow provides a stronger balance of fidelity and perceptual quality on this benchmark.

Taken together, these two public benchmarks support a consistent interpretation of the strengths of RareFlow as a cross-sensor super-resolution model. The strongest performance of RareFlow appears on the RTS benchmark, where object-centric semantic guidance is available and clearly meaningful. However, RareFlow remains strong even when the text prior may or may not present and LR-conditioned observation control carries most of the burden. In such cases, RareFlow still outperforms other model architectures in the cross-sensor super-resolution task, demonstrating the strength of the RareFlow architecture and its generalizability across different cross-sensor super-resolution scenarios in remote sensing applications.
\section{Conclusions}

This paper introduced RareFlow, a semantic-guided generative AI framework for cross-sensor remote-sensing super-resolution. RareFlow addresses the challenging problem of reconstructing 2 m Maxar-like imagery from 10 m Sentinel-2 inputs under sensor mismatch, temporal inconsistency, limited paired data, and rare geomorphic targets. RareFlow combines a frozen flow-matching generative AI backbone with LR-conditioned ControlNet residuals, learned block-wise gates, semantic guidance, and a consistency-guided training objective to improve target-sensor realism while preserving LR scene structure.

Experiments on the RTS benchmark show that RareFlow improves the balance between structural fidelity, perceptual quality, spectral consistency, and distributional alignment. It achieves the best result on six of eight main metrics and is also preferred in expert evaluation for visual interpretation of thaw-slump features. Ablation studies confirm the complementary roles of ControlNet conditioning, semantic guidance, consistency guidance, and learned gating. Additional results on SEN2NAIP and BreizhSR further suggest that the framework can generalize to other cross-sensor super-resolution settings.

The current implementation is evaluated mainly on RGB reconstructions. Future work will extend the framework to incorporate multi-temporal observations and multispectral bands, alongside explicit uncertainty calibration to support downstream tasks like automated change detection and rare-feature mapping.

\section{Acknowledgments}
This research is supported in part by Google.org’s Impact Challenge for Climate Innovation Program, the National Science Foundation (Award 2120943), and NASA Earth Surface and Interior Program (Grant 80NSSC20K0491). We thank Google Cloud Platform and the Research Computing (RC) at Arizona State University (ASU) for their support in providing computing resources to support this research. 



\bibliographystyle{IEEEtran}
\bibliography{ref}

@String(CVPR= {IEEE Conf. Comput. Vis. Pattern Recog.})

@String(ICCV= {Int. Conf. Comput. Vis.})

@String(ECCV= {Eur. Conf. Comput. Vis.})

@String(CVPR  = {CVPR})

@String(ICCV  = {ICCV})

@String(ECCV  = {ECCV})

@inproceedings{ZoomLDM,
  title     = {ZoomLDM: Latent Diffusion Model for Multi-Scale Image Generation},
  author    = {Yellapragada, Srikar and Graikos, Alexandros and Triaridis, Kostas and Prasanna, Prateek and Gupta, Rajarsi and Saltz, Joel and Samaras, Dimitris},
  booktitle = {Proceedings of the IEEE/CVF Conference on Computer Vision and Pattern Recognition (CVPR)},
  pages     = {23453--23463},
  year      = {2025}
}

@article{samsr,
  title          = {Semantic-Guided Diffusion Model for Single-Step Image Super-Resolution},
  author         = {Liu, Zihang and Zhang, Zhenyu and Tang, Hao},
  journal        = {arXiv preprint arXiv:2505.07071},
  year           = {2025},
  archivePrefix  = {arXiv},
  eprint         = {2505.07071}
}

@article{SGDM,
  title   = {Semantic Guided Large Scale Factor Remote Sensing Image Super-Resolution with Generative Diffusion Prior},
  author  = {Wang, Ce and Sun, Wanjie},
  journal = {ISPRS Journal of Photogrammetry and Remote Sensing},
  volume  = {220},
  pages   = {125--138},
  year    = {2025}
}

@article{donike2025trustworthy,
  title   = {Trustworthy Super-Resolution of Multispectral Sentinel-2 Imagery with Latent Diffusion},
  author  = {Donike, Simon and Aybar, Cesar and G{\'o}mez-Chova, Luis and Kalaitzis, Freddie},
  journal = {IEEE Journal of Selected Topics in Applied Earth Observations and Remote Sensing},
  year    = {2025}
}

@article{wan2024controlsr,
  title          = {{ControlSR}: Taming Diffusion Models for Consistent Real-World Image Super Resolution},
  author         = {Wan, Yuhao and Jiang, Peng-Tao and Hou, Qibin and Zhang, Hao and Chen, Jinwei and Cheng, Ming-Ming and Li, Bo},
  journal        = {arXiv preprint arXiv:2410.14279},
  year           = {2024},
  archivePrefix  = {arXiv},
  eprint         = {2410.14279}
}

@inproceedings{okabayashi2024cross,
  title={Cross-sensor super-resolution of irregularly sampled Sentinel-2 time series},
  author={Okabayashi, Aimi and Audebert, Nicolas and Donike, Simon and Pelletier, Charlotte},
  booktitle={Proceedings of the IEEE/CVF Conference on Computer Vision and Pattern Recognition},
  pages={502--511},
  year={2024}
}

@inproceedings{qu2024xpsr,
  title     = {{XPSR}: Cross-Modal Priors for Diffusion-Based Image Super-Resolution},
  author    = {Qu, Yunpeng and Yuan, Kun and Zhao, Kai and Xie, Qizhi and Hao, Jinhua and Sun, Ming and Zhou, Chao},
  booktitle = {European Conference on Computer Vision (ECCV)},
  pages     = {285--303},
  year      = {2024},
  publisher = {Springer Nature Switzerland}
}

@article{wang2024exploiting,
  title   = {Exploiting Diffusion Prior for Real-World Image Super-Resolution},
  author  = {Wang, Jianyi and Yue, Zongsheng and Zhou, Shangchen and Chan, Kelvin CK and Loy, Chen Change},
  journal = {International Journal of Computer Vision},
  volume  = {132},
  number  = {12},
  pages   = {5929--5949},
  year    = {2024}
}

@inproceedings{wu2024seesr,
  title     = {SeeSR: Towards Semantics-Aware Real-World Image Super-Resolution},
  author    = {Wu, Rongyuan and Yang, Tao and Sun, Lingchen and Zhang, Zhengqiang and Li, Shuai and Zhang, Lei},
  booktitle = {Proceedings of the IEEE/CVF Conference on Computer Vision and Pattern Recognition (CVPR)},
  pages     = {25456--25467},
  year      = {2024}
}

@inproceedings{yang2024pixelaware,
  title     = {Pixel-Aware Stable Diffusion for Realistic Image Super-Resolution and Personalized Stylization},
  author    = {Yang, Tao and Wu, Rongyuan and Ren, Peiran and Xie, Xuansong and Zhang, Lei},
  booktitle = {European Conference on Computer Vision (ECCV)},
  pages     = {74--91},
  year      = {2024},
  publisher = {Springer Nature Switzerland}
}

@inproceedings{blau2018perception,
  title        = {The Perception-Distortion Tradeoff},
  author       = {Blau, Yochai and Michaeli, Tomer},
  booktitle    = {IEEE/CVF Conference on Computer Vision and Pattern Recognition (CVPR)},
  pages        = {6228--6237},
  year         = {2018},
  doi          = {10.1109/CVPR.2018.00652}
}

@article{gascon2017sentinel2calib,
  title        = {Copernicus Sentinel-2A Calibration and Products Validation Status},
  author       = {Gascon, Ferran and Bouzinac, Catherine and Th{\'e}paut, Olivier and Jung, Mathieu and Francesconi, Benjamin and Lonjou, Vincent and Lafrance, Bruno and Mass{\'e}ra, St{\'e}phane and Gaudel-Vacaresse, Aurore and Languille, Fr{\'e}d{\'e}ric and Alhammoud, Bahjat and Viallefont, Fran{\c{c}}oise and Pflug, Bringfried and Bieniarz, Jakub and Clerc, St{\'e}phane and Pessiot, Laetitia and Tr{\'e}d{\'e}, Laurent and Cadau, Edoardo and De Bonis, Roberto and Isola, Claudio and Martimort, Philippe and Fernandez, Victor},
  journal      = {Remote Sensing},
  volume       = {9},
  number       = {6},
  pages        = {584},
  year         = {2017},
  doi          = {10.3390/rs9060584}
}

@misc{openai2025gpt5,
  author       = {OpenAI},
  title        = {GPT-5 Model and OpenAI API},
  year         = {2025},
  howpublished = {\url{https://platform.openai.com/docs}},
  note         = {Accessed: 2025-11-13}
}

@article{lipman2022flow,
  title={Flow matching for generative modeling},
  author={Lipman, Yaron and Chen, Ricky TQ and Ben-Hamu, Heli and Nickel, Maximilian and Le, Matt},
  journal={arXiv preprint arXiv:2210.02747},
  year={2022}
}

@inproceedings{zavadski2024controlnet,
  title={Controlnet-xs: Rethinking the control of text-to-image diffusion models as feedback-control systems},
  author={Zavadski, Denis and Feiden, Johann-Friedrich and Rother, Carsten},
  booktitle={European Conference on Computer Vision},
  pages={343--362},
  year={2024},
  organization={Springer}
}

@article{Barth2025,
  title={Rapid changes in retrogressive thaw slump dynamics in the Russian High Arctic based on very high-resolution remote sensing},
  author={Barth, S and Nitze, I and Juhls, B and Runge, Alexandra and Grosse, G},
  journal={Geophysical Research Letters},
  volume={52},
  number={7},
  pages={e2024GL113022},
  year={2025},
  publisher={Wiley Online Library}
}

@inproceedings{zhang2023adding,
  title={Adding conditional control to text-to-image diffusion models},
  author={Zhang, Lvmin and Rao, Anyi and Agrawala, Maneesh},
  booktitle={Proceedings of the IEEE/CVF international conference on computer vision},
  pages={3836--3847},
  year={2023}
}

@article{xiao2024semantic,
  title={Semantic segmentation prior for diffusion-based real-world super-resolution},
  author={Xiao, Jiahua and Zhang, Jiawei and Zou, Dongqing and Zhang, Xiaodan and Ren, Jimmy and Wei, Xing},
  journal={arXiv preprint arXiv:2412.02960},
  year={2024}
}

@inproceedings{sftgan,
  title={Recovering realistic texture in image super-resolution by deep spatial feature transform},
  author={Wang, Xintao and Yu, Ke and Dong, Chao and Loy, Chen Change},
  booktitle={Proceedings of the IEEE conference on computer vision and pattern recognition},
  pages={606--615},
  year={2018}
}

@inproceedings{Peebles2023DiT,
  title={Scalable Diffusion Models with Transformers},
  author={Peebles, William and Xie, Saining},
  booktitle={ICCV},
  year={2023},
  eprint={2212.09748}
}

@inproceedings{Jiang2021FFL,
  title={Focal Frequency Loss for Image Reconstruction and Synthesis},
  author={Jiang, Liming and Dai, Bo and Wu, Wayne and Loy, Chen Change},
  booktitle={ICCV},
  year={2021},
  eprint={2012.12821}
}

@inproceedings{Fuoli2021FourierSR,
  title={Fourier Space Losses for Efficient Perceptual Image Super-Resolution},
  author={Fuoli, Dario and Van Gool, Luc and Timofte, Radu},
  booktitle={ICCV},
  year={2021},
  eprint={2106.00783}
}

@article{yang2023mapping,
  title={Mapping retrogressive thaw slumps using deep neural networks},
  author={Yang, Yili and Rogers, Brendan M and Fiske, Greg and Watts, Jennifer and Potter, Stefano and Windholz, Tiffany and Mullen, Andrew and Nitze, Ingmar and Natali, Susan M},
  journal={Remote Sensing of Environment},
  volume={288},
  pages={113495},
  year={2023},
  publisher={Elsevier}
}

@article{wang2022review,
  title={A review of image super-resolution approaches based on deep learning and applications in remote sensing},
  author={Wang, Xuan and Yi, Jinglei and Guo, Jian and Song, Yongchao and Lyu, Jun and Xu, Jindong and Yan, Weiqing and Zhao, Jindong and Cai, Qing and Min, Haigen},
  journal={Remote Sensing},
  volume={14},
  number={21},
  pages={5423},
  year={2022},
  publisher={MDPI}
}

@inproceedings{esser2024scaling,
  title={Scaling rectified flow transformers for high-resolution image synthesis},
  author={Esser, Patrick and Kulal, Sumith and Blattmann, Andreas and Entezari, Rahim and M{\"u}ller, Jonas and Saini, Harry and Levi, Yam and Lorenz, Dominik and Sauer, Axel and Boesel, Frederic and others},
  booktitle={Forty-first international conference on machine learning},
  year={2024}
}

@article{moser2024diffusion,
  title={Diffusion models, image super-resolution, and everything: A survey},
  author={Moser, Brian B and Shanbhag, Arundhati S and Raue, Federico and Frolov, Stanislav and Palacio, Sebastian and Dengel, Andreas},
  journal={IEEE Transactions on Neural Networks and Learning Systems},
  year={2024},
  publisher={IEEE}
}

@article{ssim,
  title={Image quality assessment: from error visibility to structural similarity},
  author={Wang, Zhou and Bovik, Alan C and Sheikh, Hamid R and Simoncelli, Eero P},
  journal={IEEE transactions on image processing},
  volume={13},
  number={4},
  pages={600--612},
  year={2004},
  publisher={IEEE}
}

@inproceedings{lpips,
  title={The unreasonable effectiveness of deep features as a perceptual metric},
  author={Zhang, Richard and Isola, Phillip and Efros, Alexei A and Shechtman, Eli and Wang, Oliver},
  booktitle={Proceedings of the IEEE conference on computer vision and pattern recognition},
  pages={586--595},
  year={2018}
}

@inproceedings{maniqa,
  title={Maniqa: Multi-dimension attention network for no-reference image quality assessment},
  author={Yang, Sidi and Wu, Tianhe and Shi, Shuwei and Lao, Shanshan and Gong, Yuan and Cao, Mingdeng and Wang, Jiahao and Yang, Yujiu},
  booktitle={Proceedings of the IEEE/CVF conference on computer vision and pattern recognition},
  pages={1191--1200},
  year={2022}
}

@article{niqe,
  title={Making a “completely blind” image quality analyzer},
  author={Mittal, Anish and Soundararajan, Rajiv and Bovik, Alan C},
  journal={IEEE Signal processing letters},
  volume={20},
  number={3},
  pages={209--212},
  year={2012},
  publisher={IEEE}
}

@inproceedings{psnr,
  title={Image quality metrics: PSNR vs. SSIM},
  author={Hore, Alain and Ziou, Djemel},
  booktitle={2010 20th international conference on pattern recognition},
  pages={2366--2369},
  year={2010},
  organization={IEEE}
}

@article{fid,
  title={Gans trained by a two time-scale update rule converge to a local nash equilibrium},
  author={Heusel, Martin and Ramsauer, Hubert and Unterthiner, Thomas and Nessler, Bernhard and Hochreiter, Sepp},
  journal={Advances in neural information processing systems},
  volume={30},
  year={2017}
}

@misc{ESA_Sentinel2,
  title        = {Sentinel-2: Colour vision for Copernicus},
  author       = {{European Space Agency}},
  howpublished = {\url{https://www.esa.int/Applications/Observing_the_Earth/Copernicus/Sentinel-2}},
  year         = {2025},
  note         = {Accessed 2025-11-13}
}

@misc{Maxar_WorldView3,
  title        = {WorldView-3},
  author       = {{European Space Agency}},
  howpublished = {\url{https://earth.esa.int/eogateway/missions/worldview-3}},
  year         = {2025},
  note         = {Accessed 2025-11-13}
}

@article{aybar2024sen2naip,
  title={SEN2NAIP: A large-scale dataset for Sentinel-2 Image Super-Resolution},
  author={Aybar, Cesar and Montero, David and Contreras, Julio and Donike, Simon and Kalaitzis, Freddie and G{\'o}mez-Chova, Luis},
  journal={Scientific Data},
  volume={11},
  number={1},
  pages={1389},
  year={2024},
  publisher={Nature Publishing Group UK London}
}

@article{caldareri2024shoreline,
  title={On the shoreline monitoring via earth observation: An isoradiometric method},
  author={Caldareri, Francesco and Sulli, Attilio and Parrino, Nicolo' and Dardanelli, Gino and Todaro, Simona and Maltese, Antonino},
  journal={Remote Sensing of Environment},
  volume={311},
  pages={114286},
  year={2024},
  publisher={Elsevier}
}

@article{turner2015free,
  title={Free and open-access satellite data are key to biodiversity conservation},
  author={Turner, Woody and Rondinini, Carlo and Pettorelli, Nathalie and Mora, Brice and Leidner, Allison K and Szantoi, Zoltan and Buchanan, Graeme and Dech, Stefan and Dwyer, John and Herold, Martin and others},
  journal={Biological conservation},
  volume={182},
  pages={173--176},
  year={2015},
  publisher={Elsevier}
}

@article{liu2021research,
  title={Research on super-resolution reconstruction of remote sensing images: A comprehensive review},
  author={Liu, Hui and Qian, Yurong and Zhong, Xiwu and Chen, Long and Yang, Guangqi},
  journal={Optical Engineering},
  volume={60},
  number={10},
  pages={100901--100901},
  year={2021},
  publisher={Society of Photo-Optical Instrumentation Engineers}
}

@article{dong2015image,
  title={Image super-resolution using deep convolutional networks},
  author={Dong, Chao and Loy, Chen Change and He, Kaiming and Tang, Xiaoou},
  journal={IEEE transactions on pattern analysis and machine intelligence},
  volume={38},
  number={2},
  pages={295--307},
  year={2015},
  publisher={IEEE}
}

@article{lanaras2018super,
  title={Super-resolution of Sentinel-2 images: Learning a globally applicable deep neural network},
  author={Lanaras, Charis and Bioucas-Dias, Jos{\'e} and Galliani, Silvano and Baltsavias, Emmanuel and Schindler, Konrad},
  journal={ISPRS Journal of Photogrammetry and Remote Sensing},
  volume={146},
  pages={305--319},
  year={2018},
  publisher={Elsevier}
}

@inproceedings{liang2021swinir,
  title={Swinir: Image restoration using swin transformer},
  author={Liang, Jingyun and Cao, Jiezhang and Sun, Guolei and Zhang, Kai and Van Gool, Luc and Timofte, Radu},
  booktitle={Proceedings of the IEEE/CVF international conference on computer vision},
  pages={1833--1844},
  year={2021}
}

@article{kowaleczko2023real,
  title={A real-world benchmark for Sentinel-2 multi-image super-resolution},
  author={Kowaleczko, Pawel and Tarasiewicz, Tomasz and Ziaja, Maciej and Kostrzewa, Daniel and Nalepa, Jakub and Rokita, Przemyslaw and Kawulok, Michal},
  journal={Scientific Data},
  volume={10},
  number={1},
  pages={644},
  year={2023},
  publisher={Nature Publishing Group UK London}
}

@inproceedings{dong2024building,
  title={Building bridges across spatial and temporal resolutions: Reference-based super-resolution via change priors and conditional diffusion model},
  author={Dong, Runmin and Yuan, Shuai and Luo, Bin and Chen, Mengxuan and Zhang, Jinxiao and Zhang, Lixian and Li, Weijia and Zheng, Juepeng and Fu, Haohuan},
  booktitle={Proceedings of the ieee/cvf conference on computer vision and pattern recognition},
  pages={27684--27694},
  year={2024}
}

@article{michel2025revisiting,
  title={Revisiting remote sensing cross-sensor Single Image Super-Resolution: the overlooked impact of geometric and radiometric distortion},
  author={Michel, Julien and Kalinicheva, Ekaterina and Inglada, Jordi},
  journal={IEEE Transactions on Geoscience and Remote Sensing},
  year={2025},
  publisher={IEEE}
}

@article{aybar2026radiometrically,
  title={A radiometrically and spatially consistent super-resolution framework for Sentinel-2},
  author={Aybar, Cesar and Contreras, Julio and Donike, Simon and Portal{\'e}s-Juli{\`a}, Enrique and Mateo-Garc{\'\i}a, Gonzalo and G{\'o}mez-Chova, Luis},
  journal={Remote Sensing of Environment},
  volume={334},
  pages={115222},
  year={2026},
  publisher={Elsevier}
}

@inproceedings{zhang2025uncertainty,
  title={Uncertainty-guided perturbation for image super-resolution diffusion model},
  author={Zhang, Leheng and You, Weiyi and Shi, Kexuan and Gu, Shuhang},
  booktitle={Proceedings of the Computer Vision and Pattern Recognition Conference},
  pages={17980--17989},
  year={2025}
}

@article{qiu2023cross,
  title={Cross-sensor remote sensing imagery super-resolution via an edge-guided attention-based network},
  author={Qiu, Zhonghang and Shen, Huanfeng and Yue, Linwei and Zheng, Guizhou},
  journal={ISPRS Journal of Photogrammetry and Remote Sensing},
  volume={199},
  pages={226--241},
  year={2023},
  publisher={Elsevier}
}

@article{li2023convformersr,
  title={ConvFormerSR: Fusing transformers and convolutional neural networks for cross-sensor remote sensing imagery super-resolution},
  author={Li, Junjie and Meng, Yizhuo and Tao, Chongxin and Zhang, Zhen and Yang, Xining and Wang, Zhe and Wang, Xi and Li, Linyi and Zhang, Wen},
  journal={IEEE Transactions on Geoscience and Remote Sensing},
  volume={62},
  pages={1--15},
  year={2023},
  publisher={IEEE}
}

@article{latte2020planetscope,
  title={PlanetScope radiometric normalization and sentinel-2 super-resolution (2.5 m): A straightforward spectral-spatial fusion of multi-satellite multi-sensor images using residual convolutional neural networks},
  author={Latte, Nicolas and Lejeune, Philippe},
  journal={Remote Sensing},
  volume={12},
  number={15},
  pages={2366},
  year={2020},
  publisher={MDPI}
}

@article{qi2026advancing,
  title={Advancing image super-resolution techniques in remote sensing: A comprehensive survey},
  author={Qi, Yunliang and Lou, Meng and Liu, Yimin and Li, Lu and Yang, Zhen and Nie, Wen},
  journal={ISPRS Journal of Photogrammetry and Remote Sensing},
  volume={231},
  pages={68--100},
  year={2026},
  publisher={Elsevier}
}

@inproceedings{le2024detecting,
  title={Detecting out-of-distribution earth observation images with diffusion models},
  author={Le Bellier, Georges and Audebert, Nicolas},
  booktitle={Proceedings of the IEEE/CVF Conference on Computer Vision and Pattern Recognition},
  pages={481--491},
  year={2024}
}

@inproceedings{ledig2017photo,
  title={Photo-realistic single image super-resolution using a generative adversarial network},
  author={Ledig, Christian and Theis, Lucas and Husz{\'a}r, Ferenc and Caballero, Jose and Cunningham, Andrew and Acosta, Alejandro and Aitken, Andrew and Tejani, Alykhan and Totz, Johannes and Wang, Zehan and others},
  booktitle={Proceedings of the IEEE conference on computer vision and pattern recognition},
  pages={4681--4690},
  year={2017}
}

@article{saharia2022image,
  title={Image super-resolution via iterative refinement},
  author={Saharia, Chitwan and Ho, Jonathan and Chan, William and Salimans, Tim and Fleet, David J and Norouzi, Mohammad},
  journal={IEEE transactions on pattern analysis and machine intelligence},
  volume={45},
  number={4},
  pages={4713--4726},
  year={2022},
  publisher={IEEE}
}

@article{lu2025multi,
  title={Multi-level Priors-Guided Diffusion-based Remote Sensing Image Super-Resolution},
  author={Lu, Lijing and Huang, Zhou and Bao, Yi and Wan, Lin and Li, Zhihang},
  journal={ISPRS Journal of Photogrammetry and Remote Sensing},
  volume={228},
  pages={756--770},
  year={2025},
  publisher={Elsevier}
}

@inproceedings{chen2025adversarial,
  title={Adversarial diffusion compression for real-world image super-resolution},
  author={Chen, Bin and Li, Gehui and Wu, Rongyuan and Zhang, Xindong and Chen, Jie and Zhang, Jian and Zhang, Lei},
  booktitle={Proceedings of the IEEE/CVF conference on computer vision and pattern recognition},
  pages={28208--28220},
  year={2025}
}

@inproceedings{gandikota2024text,
  title={Text-guided explorable image super-resolution},
  author={Gandikota, Kanchana Vaishnavi and Chandramouli, Paramanand},
  booktitle={Proceedings of the IEEE/CVF conference on computer vision and pattern recognition},
  pages={25900--25911},
  year={2024}
}

@article{hayes2007krippendorff,
  title={Answering the Call for a Standard Reliability Measure for Coding Data},
  author={Hayes, Andrew F. and Krippendorff, Klaus},
  journal={Communication Methods and Measures},
  volume={1},
  number={1},
  pages={77--89},
  year={2007},
  doi={10.1080/19312450709336664}
}

@article{kendall1939rankings,
  title={The Problem of $m$ Rankings},
  author={Kendall, M. G. and Smith, B. Babington},
  journal={The Annals of Mathematical Statistics},
  volume={10},
  number={3},
  pages={275--287},
  year={1939},
  doi={10.1214/aoms/1177732186}
}

@article{spearman1904association,
  title={The Proof and Measurement of Association Between Two Things},
  author={Spearman, C.},
  journal={The American Journal of Psychology},
  volume={15},
  number={1},
  pages={72--101},
  year={1904},
  doi={10.2307/1412159}
}

@article{liljedahl2016pan,
  title={Pan-Arctic ice-wedge degradation in warming permafrost and its influence on tundra hydrology},
  author={Liljedahl, Anna K and Boike, Julia and Daanen, Ronald P and Fedorov, Alexander N and Frost, Gerald V and Grosse, Guido and Hinzman, Larry D and Iijma, Yoshihiro and Jorgenson, Janet C and Matveyeva, Nadya and others},
  journal={Nature Geoscience},
  volume={9},
  number={4},
  pages={312--318},
  year={2016},
  publisher={Nature Publishing Group UK London}
}

@article{nitze2025darts,
  title={DARTS: Multi-year database of AI-detected retrogressive thaw slumps in the circum-arctic permafrost region},
  author={Nitze, Ingmar and Heidler, Konrad and Nesterova, Nina and K{\"u}pper, Jonas and Sch{\"u}tt, Emma and H{\"o}lzer, Tobias and Barth, Sophia and Lara, Mark J and Liljedahl, Anna K and Grosse, Guido},
  journal={Scientific Data},
  volume={12},
  number={1},
  pages={1512},
  year={2025},
  publisher={Nature Publishing Group UK London}
}

@article{casagli2023landslide,
  title={Landslide detection, monitoring and prediction with remote-sensing techniques},
  author={Casagli, Nicola and Intrieri, Emanuele and Tofani, Veronica and Gigli, Giovanni and Raspini, Federico},
  journal={Nature Reviews Earth \& Environment},
  volume={4},
  number={1},
  pages={51--64},
  year={2023},
  publisher={Nature Publishing Group UK London}
}

\end{document}